\documentclass[11pt]{article}
\usepackage{paralist} % to compact itemize, enumerate, description, like asparaenum
\usepackage{emnlp2021}

\usepackage{times}
\usepackage{latexsym}
\usepackage[T1]{fontenc}
\usepackage[utf8]{inputenc}

\usepackage{microtype}

\usepackage{url}
\usepackage{amssymb}
\usepackage{hyperref}
\usepackage{graphicx}
\usepackage{booktabs}
\usepackage{multirow}
\usepackage{hyperref}
\usepackage{subcaption}

\usepackage[most]{tcolorbox}

\usepackage[title]{appendix}

\usepackage{xargs}
\usepackage[prependcaption,textsize=tiny]{todonotes}
\newcommandx{\VGn}[2][1=]{\todo[linecolor=blue,backgroundcolor=blue!25,bordercolor=blue,#1]{#2}}
\newcommandx{\NAn}[2][1=]{\todo[linecolor=red,backgroundcolor=red!25,bordercolor=red,#1]{#2}}
\newcommandx{\AKn}[2][1=]{\todo[linecolor=green,backgroundcolor=green!25,bordercolor=green,#1]{#2}}
\newcommandx{\DNn}[2][1=]{\todo[linecolor=gray,backgroundcolor=gray!25,bordercolor=gray,#1]{OK: #2}}

\newcommandx{\Addition}[1]{#1}
\title{Uncovering the Limits of Text-based Emotion Detection}
\author{Nurudin Alvarez-Gonzalez\textsuperscript{1,*}, Andreas Kaltenbrunner\textsuperscript{1,2}, 
Vicen\c{c} G\'omez\textsuperscript{1}\\
{\normalfont
nuralgon@gmail.com,
kaltenbrunner@gmail.com, 
vicen.gomez@upf.edu}\\
\textsuperscript{1}Universitat Pompeu Fabra. Barcelona, Spain.\\
\textsuperscript{2}ISI Foundation. Turin, Italy.}

\begin{document}

\maketitle

\let\thefootnote\relax\footnotetext{Accepted for publication in Findings of EMNLP 2021.}

% \Addition{\textbf{Other paper on Vent. Main points:}}

% \Addition{1. Our work does not limit the analysis of emotions to categories that overlap with established emotion taxonomies, such as recent works on the Vent dataset~\cite{malko-etal-2021-demonstrating}.}

% \Addition{2. Related works focusing only on emotion categories~\cite{malko-etal-2021-demonstrating} found that multi-class models 75\% of the errors in their model could be attributed to ambiguity, in which both the original Vent label and the output of the models would be conceivable emotional signals contained in the text.} 

%\begin{itemize}
%    \item Cite new paper on Vent, compare approaches. Ref: [1] Malko et al. Demonstrating the Reliability of Self-Annotated Emotion Data. CLPsych | NAACL (2021)
%    \item Clean up paper and remove 'hanging' lines.
%\end{itemize}

\begin{abstract}
%Identifying emotions from text is crucial for a variety of real world tasks. We consider the two largest now-available corpora for emotion classification: GoEmotions and Vent. The first contains multiple emotion annotations extracted from 58k Reddit comments, while the second contains 33 million messages with a single emotion chosen by the writer. We design a benchmark and evaluate several feature spaces and learning algorithms. Our results on GoEmotions outperform previous strong baselines using BERT. Through an experiment with human participans, we also analyse the differences between how writers express emotions and how readers perceive them. Our results suggest that identifying emotions that writers express might be harder than detecting emotions readers perceive. Our learned models are open for researchers to explore through a public interface.

Identifying emotions from text is crucial for a variety of real world tasks. We consider the two largest now-available corpora for emotion classification: GoEmotions, with 58k messages labelled by readers, and Vent, with 33M writer-labelled messages. We design a benchmark and evaluate several feature spaces and learning algorithms, including two simple yet novel models on top of BERT that outperform previous strong baselines on GoEmotions. Through an experiment with human participants, we also analyze the differences between how writers express emotions and how readers perceive them. Our results suggest that emotions expressed by writers are harder to identify than emotions that readers perceive. We share a public web interface for researchers to explore our models.

%Identifying emotions from text is crucial for a variety of real world tasks. In this work, we study the limits of text-based emotion detection by analysing datasets, models, and label perspectives. We design a benchmark spanning both established and novel yet simple modelling techniques on two recent large scale datasets, and investigate the difference in emotional labelling between authors of social media posts and readers engaging with their content. Our models outperform previous baselines and humans, according to annotations that we collect. Using our annotations, we study the differences between how writers express emotions and how readers perceive them. Our results suggest that identifying emotions that writers express might be harder than detecting emotions readers perceive.

\end{abstract}

\section{Introduction}
\label{sec:Intro}

%Emotional aspects of communication play a role in expressing subjective states of mind and defining our social interactions. Emotions underpin our language when we share humor or grief, when we signal isolation or group membership, or when we agree or disagree with political discourses.

Identifying emotional signals is key to a series of downstream tasks. For instance, emotion detection is necessary for empathetic chat-bots that can respond to the emotional needs of their users~\cite{fung2016}. Distinguishing emotional content is required to study viral~\cite{guerini2015}, educational~\cite{ortigosa2014}, political~\cite{mohammad2015}, or incendiary~\cite{brassard-gourdeau2019} interactions on social media. Capturing the evolution of user-provided emotional content can help prevent harassment~\cite{agrawal2018} or develop early indicators for depression~\cite{husseini2018,jimir20}. Facial expressions~\cite{li2020}, speech~\cite{elayadi2011}, body movements~\cite{noroozi2018} and text~\cite{poria2019} are sources from which emotions may be automatically extracted. 

%Emotion analysis is typically harder than sentiment analysis, which characterizes text in terms of polarity (positive, negative or neutral)\NAn{Not sure if 'harder' is the right term}. In contrast, emotion analysis involves a larger set of classes, and is influenced by aspects such as ambiguity, misunderstandings, irony or sarcasm~\cite{mohammad2021sentiment,chauhan2020}. Much of the recent progress in emotion analysis has been enabled by two main factors: the success of pre-trained language models, such as Bidirectional Encoder Representations from Transformers (BERT)~\cite{devlin2019}, and the increasing availability of high-quality large-scale annotated datasets.

Emotion analysis contrasts sentiment analysis, which characterizes text in terms of polarity (positive, negative or neutral), by involving a larger set of classes, often influenced by aspects such as ambiguity, misunderstandings, irony or sarcasm~\cite{mohammad2021sentiment,chauhan2020}. Recent progress in the field has been enabled by the success of pre-trained language models, such as Bidirectional Encoder Representations from Transformers (BERT)~\cite{devlin2019}, and the release of high-quality large-scale annotated datasets.

In this work, we analyze the limits of text-based emotion detection on the two largest now-available corpora: GoEmotions~\cite{demszky2020} and Vent~\cite{lykousas2019}. GoEmotions contains 58k Reddit comments tagged with possibly multiple labels out 28 emotions, annotated by third-person \emph{readers}. The raw Vent corpus~\cite{lykousas2019} includes 33M messages tagged with one out of 705 emotions by their original first-person \emph{writers}. The unprecedented volume of these datasets makes them suitable to study textual emotion detection at scale from different perspectives.

%In this work, we analise the current limits of text-based emotion detection on the two largest available corpora at the moment: GoEmotions~\cite{demszky2020} and Vent~\cite{lykousas2019}. GoEmotions is a corpus comprising 58k Reddit comments with, possibly multiple, annotated emotions from a set of 27 different emotion labels plus a neutral category. The raw Vent dataset~\cite{lykousas2019} contains 33M messages, each of them containing a single emotion selected by its writer from 705 possible emotions grouped into 63 categories. The unprecedented volume of these datasets makes them suitable to analise textual emotion detection at scale from different perspectives.

%\footnote{Anonymized interface to our predictive models at:\\
%\noindent{\href://emotionui-anon.s3-website-us-west-1.amazonaws.com/}{http://emotionui-anon.s3-website-us-west-1.amazonaws.com}}.

%In this paper, we analise the limits of text-based emotion detection on the two largest available corpora at the moment: GoEmotions~\cite{demszky2020} and Vent~\cite{lykousas2019}. GoEmotions is a manually annotated corpus comprising 58k Reddit comments with 27 different emotion labels plus a neutral category. The raw Vent dataset~\cite{lykousas2019} contains 33M messages, each of them containing the user's self-annotated emotion chosen from 705 possible emotions grouped into 63 emotion categories. The unprecedented volume of these datasets makes them particularly suitable to analise textual emotion detection at scale from different perspectives.

We first focus on the predictive performance of several feature spaces and learning algorithms. Our benchmark includes simple models that outperform previous baselines, including a strong BERT baseline on GoEmotions. We find that statistical methods such as TF-IDF outperform more complex word-level embeddings such as FastText. Finally, we release a web interface showcasing our models.

Second, we analyze the hierarchical structure of the label space for models trained on Vent. Our models capture the cluster structure defined by emotion categories to a large extent, despite not explicitly observing emotion categories during training.

Finally, we design an experiment with human participants to evaluate our model and the differences between emotions provided by writers and those perceived by participants (readers). Our models outperform readers at predicting emotions intended by writers, and they also predict the emotions annotated by readers even more accurately, with important implications for emotion analysis.

\section{Related Work}
\label{sec:RelatedWork}

We discuss three domains of related work: the existing taxonomies to represent emotions, the corpora used to build and evaluate emotion detection systems, and the NLP approaches that may be used to implement text-based emotion detection systems. 

\subsection{Emotion Taxonomies}

The landscape of human emotion has been represented by several different taxonomies and approaches. Ekman~\cite{ekman1971} proposed 6 basic emotions expressed through facial expressions across cultures: \emph{joy, sadness, anger, surprise, disgust and fear}. Independently, Plutchik~\cite{plutchik1980} introduced a similar taxonomy that added \emph{anticipation} and \emph{trust}, characterizing emotions through his Wheel of Emotion~\cite{plutchik1980}.%\footnote{\url{https://en.wikipedia.org/wiki/Robert_Plutchik\#/media/File:Plutchik-wheel.svg}}. %Recent advances in Psychology have proposed 
Finer-grained emotion taxonomies have been recently proposed, capturing high-dimensional relationships of up to 600 different emotions, clustering emotion concepts using machine learning techniques~\cite{cowen2019}. These taxonomies show the multifaceted nature of emotions across cultures in vocalization~\cite{cowen2018}, music~\cite{cowen2020}, or facial expressions~\cite{cowen2019b}. Beyond these discrete categorizations, other models such as the affective-circumplex model of emotions~\cite{rusell2003}, have captured emotions as proportions on three dimensions rather than discrete categories: Valence (positive or negative), Activation (active or passive), and Dominance (dominant or submissive). Finally, a small number of works extend these taxonomies to include the perspective of senders and receivers of emotional communication~\cite{Ptaszynski2009,mohammad2013b,buechel2017,buechel2017b}.

In this work, we focus on categorical approaches with recent emotional taxonomies covering a rich spectrum of emotions from the perspectives of senders and receivers. 
%taxonomy similar to Ekman's

%\begin{figure}[ht]
%    \centering
%    \includegraphics[width=1.0\linewidth]{diagrams/Plutchik-wheel.pdf}
%    \caption[Plutchik's Wheel of Emotion.]{Plutchik's Wheel of Emotion. Similar emotions are adjacent to one another and contrasting ones are found on opposite ends of the wheel. Graph depicting the wheel taken from Wikipedia~\footnotemark.}
%    \label{fig:WheelEmotion}
%\end{figure}

\subsection{Emotion Detection Text Corpora}

To build Emotion Detection systems, practitioners require text data containing emotional signals. Early works like SentiStrength~\cite{thelwall2010} and ANEW~\cite{nielsen2011} used lexical associations for sentiment analysis, capturing whether text was positive, negative or neutral and to which degree. Lexical approaches can be used in rule-based systems, where words contribute to a sentiment or emotion signal. Sophisticated rule-based models like VADER~\cite{hutto2014} rely on human-annotated word sentiments, alongside with slang, modifiers, emphasis or punctuation. Other approaches went beyond polarity: LIWC~\cite{tausczik2010} presented labelled dictionaries mapping words to their emotional and polarity probabilities. Likewise, EmoLex~\cite{mohammad2013} crowd-sourced a word-emotion association lexicon labelling words in terms of sentiment and emotion following Plutchik's taxonomy. 

Another approach is to treat emotion detection as a supervised learning problem, with corpora including emotional information varying in size, scope and labelling approach~\cite{bostan2018}. Early datasets such as Affective Text~\cite{strapparava2007} were small, with 1,250 headlines labelled for valence and Ekman's emotions. To reduce data acquisition efforts, some works have explored corpora mined from emotion-rich environments such as social networks. For instance, Crowdflower's emotion dataset~\cite{crowdflower2016} labelled 40K tweets expanding Ekman's emotions with \emph{enthusiasm, fun, hate, neutral, love, boredom, relief} and \emph{empty}. Similarly, EmoNet~\cite{abdul2017} applied distant supervision by labelling tweets using hashtags among the `circles' within Plutchik's Wheel. \Addition{Finally, in parallel to our work,~\citet{malko-etal-2021-demonstrating} used Vent data on a limited number of emotion categories overlapping with Ekman's emotions, and concluded that the self-annotated labels of Vent are indicative of emotional contents expressed in the text, supporting more detailed analyses of emotion expression.}

If emotion detection is treated as a learning problem, datasets present a trade-off between size and quality. Human-annotated datasets span thousands of samples, often targeting a small number of emotions, such as SocialNLP 2019 EmotionX challenge or Crowdflower ~\cite{strapparava2007,crowdflower2016,shmueli2019}. Approaches like EmoNet~\cite{abdul2017} can comprise millions of samples collected from social media using distantly supervised labels, allowing for larger datasets at the time of publication. However, datasets collected from social media may be private, with direct dataset sharing often being forbidden, and content routinely getting deleted, limiting
reproducibility. Additionally, labels produced by distant supervision using hashtags might not align with how humans generally perceive or express emotions across domains. 

Our work addresses these shortcomings by exploring the two largest human-annotated emotion datasets, GoEmotions and Vent, the second of which features 33M messages annotated into 705 emotions by their authors in a social network. We describe these two corpora in Sections~\ref{sec:goemo} and~\ref{sec:VentDataset}. 

%GoEmotions~\cite{demszky2020} introduced a manually annotated corpus of 58k Reddit comments with labels under 27 different emotions plus a neutral category. Likewise, the Vent dataset~\cite{lykousas2019} contains 33 million messages posted on a social network were users self-annotate the emotion of each post into 705 emotions divided in 63 emotion categories. The volume of data in GoEmotions and Vent make them suitable to capture richer emotional signals from text by applying state-of-the-art NLP techniques. In this work we focus on large-scale emotion corpora by working on the GoEmotions and Vent datasets.

\subsection{NLP Emotion Detection Models}

Approaches to Emotion Detection are often constrained by the difficulties of extracting emotional signals from small collections of labelled data~\cite{alswaidan2020,acheampong2020}, using both feature-engineered and neural models. Some feature-based models introduce emotional priors by using word-emotion associations as features for supervised classifiers~\cite{mohammad2013}. Other approaches employ statistical methods such Bag-of-Words to represent documents, placing the effort of learning emotional associations on the model~\cite{silva2014}. 

Recently, larger datasets have allowed to train neural models that outperform their traditional counterparts. For instance, EmoNet involved collecting a 1.6M tweet dataset, and training RNN-based models on the same data~\cite{abdul2017}. After the release of BERT~\cite{devlin2019}, an explosion of novel work has focused on fine-tuning transformer models to learn from scarce emotion data. For example, the top performing models on the SocialNLP 2019 EmotionX Challenge~\cite{shmueli2019} outperform the best previous existing model by a 19\% increase in micro-F1. In this direction, GoEmotions introduced a fine-tuned BERT multi-label classifier baseline with 46\% macro-F1 across 28 possible labels~\cite{demszky2020}. Modelling choices have often depended on the availability of data: rule-based~\cite{hutto2014,tausczik2010} and lexical approaches~\cite{mohammad2013} when training data was too sparse for ML-based solutions, non-neural methods as datasets scaled~\cite{silva2014} and, more recently, pre-trained neural models and transformer architectures. The outstanding results of fine-tuned transformers have driven a majority of recent work, including the SocialNLP 2019 EmotionX challenge~\cite{shmueli2019}, GoEmotions~\cite{demszky2020}, and emotion detection and sentiment analysis benchmarks~\cite{acheampong2021}. 

We explore simple Transformer-based baselines on this common ground, in which we use BERT as a representation layer and include lightweight models on top of the contextualized embeddings, outperforming previous BERT baselines on micro-F1 by 11.8\%. However, we also study non-neural and non-Transformer methods that remain popular in industry. Our contribution covers different approaches on large scale emotion datasets with rich label spaces and 58K / 9.75M sample texts. Rather than narrowly benchmarking variations of specific architectures, e.g., fine-tuning transformer language models, we work with a variety of established methods to help practitioners choose a modelling approach in terms of predictive performance and model complexity. 

\section{Emotion Detection in Text}
\label{sec:EmoDetect}

%We formulate Emotion Detection as a classification problem: given an input text, the objective is to identify most likely emotional readings of that text. A text snippet can have several emotions associated with in in cases of ambiguity, or when expressing multiple feelings at once, as seen in~\autoref{tab:EmotionExamples}.
%Our objective is to identify most likely emotional readings from text sentences.
A text snippet can have several associated emotions in cases of ambiguity, or when expressing multiple feelings at once, as seen in~\autoref{tab:EmotionExamples}. As such, we represent the problem as multiple valid labels being possible for a given snippet. %\Addition{We represent the problem , which can then be predicted by e.g. using a multi-label classification model.}

\begin{table}[!ht]
\centering
\begin{tabular}{@{}l@{\hskip 2pt}l@{}}
\toprule
\multicolumn{1}{c}{\textbf{Text}} & \textbf{Emotion} \\ \midrule
\begin{tabular}[c]{@{}l@{}}Wow. I just read the synopsis, \\ and  that’s  really what happens.\end{tabular} & Surprise \\ \midrule
\begin{tabular}[c]{@{}l@{}}What do you think? If you look \\ at my question above? Last \\ thing I should do is to say sorry?\end{tabular} & Confusion \\ \midrule
\begin{tabular}[c]{@{}l@{}}And then everyone clapped  \\ and  cheered.\end{tabular} & \begin{tabular}[c]{@{}l@{}}Joy, \\Admiration \end{tabular} \\ \bottomrule
\end{tabular}
\caption{Labelled text-emotion pairs from the GoEmotions dataset. The third example shows an instance with multiple emotional associations.}
\label{tab:EmotionExamples}
\end{table}

\subsection{Task Definition}
\label{sec:TaskDefinition}

We formulate emotion detection as a multi-label classification task. Given an input text $\mathbf{s}~\in~\mathcal{U}^*$ and a set of $N$ emotions, our task is to produce (learn) a function $\phi: \mathcal{U}^* \mapsto \mathcal{P}^{N}$ that maps $\mathbf{s}$ into independent probabilities for each emotions $y_1, y_2, ..., y_{N}$.
Treating emotion detection as a multi-label task allows us to apply the same architecture on multi-class datasets. It also allows us to account for ambiguity, even in datasets where only a single output class is expected, such as Vent. 

\section{Experimental Design}
\label{sec:ExperimentalDesign}

Here we briefly describe the GoEmotions and Vent datasets, 
our multi-label classification benchmark design, and
the representation and modelling approaches that serve as the building blocks for our emotion detectors. We include extensive details on the experimental design, hyper-parameters, additional results, and data analyzes in the Supplemental Materials.

\subsection{GoEmotions}\label{sec:goemo}

\begin{figure}[!ht]
    \centering
    \includegraphics[width=0.95\linewidth]{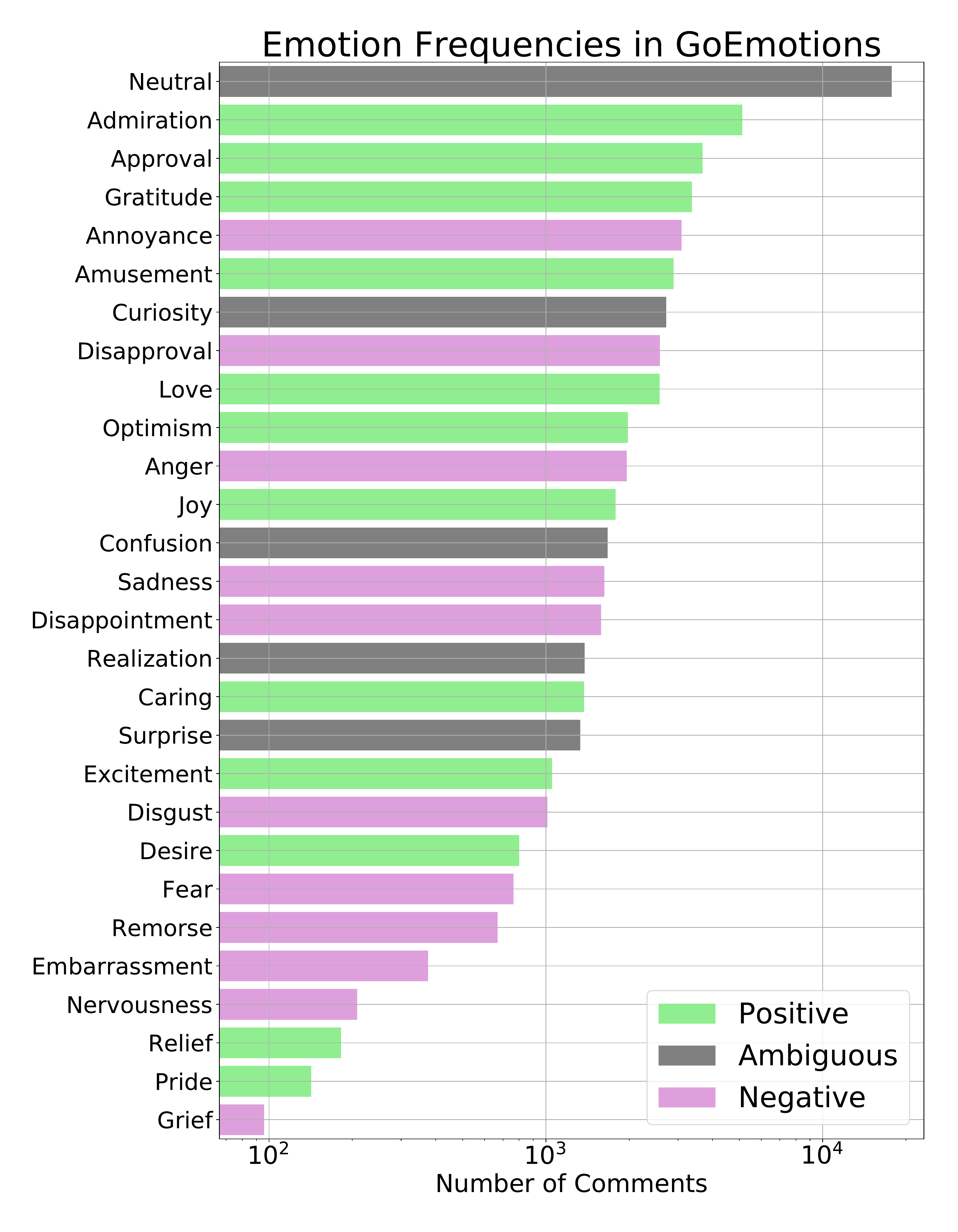}
    \caption{GoEmotions label frequencies and the three categories in~\cite{demszky2020}: Positive, Ambiguous and Negative. }
    \label{fig:GoEmotionsLabels}
\end{figure}

The GoEmotions dataset~\cite{demszky2020} contains 58,009 text snippets collected from English Reddit comments. A minimum of 3 raters labelled each snippet into multiple emotions from 27 emotion categories plus a neutral category, keeping only snippets where 2+ raters agree on one or more labels.  \autoref{fig:GoEmotionsLabels} shows the frequencies and types of each emotion in the dataset. Raters were asked to label comments as ``Neutral'' when they were not able to clearly assign an emotion to a comment. %Finally, several labels can be assigned to a specific comment. 
The dataset is accompanied with a strong baseline fine-tuned BERT classifier built on top of BERT-base~\cite{devlin2019} that predicts the 28 emotions from the contextualized embedding of the last token, achieving 0.46 macro-F1 and 0.51 micro-F1 scores. %for emotion detection. %We use these results when comparing our approach in~\autoref{tab:GoEmotionsBenchmark}, acting as a baseline to validate the design of our benchmark.

\subsection{Vent}
\label{sec:VentDataset}

The Vent dataset~\cite{lykousas2019} contains more than $33$M comments from a social network and its accompanied mobile app, predominantly in English. Each comment or ``vent'' is self-annotated by the author, which we will refer to as ``\emph{venter}'', according to their emotional state. Emotions are structured into 63 emotion categories covering 704 emotions. 

In contrast to GoEmotions where comments are labelled with ``\emph{reader}'' emotions that annotators infer from text, Vent is a ``\emph{writer}'' emotion dataset---every Vent comment is labelled with one subjective emotional label provided by its writer. The dataset is provided as-is, with minimal pre-processing anonymising user and URL references. To normalize the corpus, we \emph{(a)} remove stylistic highlighting such as italics, \emph{(b)} eliminate extraneous white space, and \emph{(c)} map user and URL references to special fixed tokens. Since the length of Vents is heterogeneous, we limit comment length to the range between 3 and 32 tokens (75-th percentile). We restrict ourselves to categories that contain at least one valid emotion, and ignore emotions marked as disabled, not used at least once per month, or whose meaning is ambiguous, e.g., those containing emoji like ``Mushy''. The filtered data contains 9.75M comments in 88 emotions that are grouped into 9 categories.  

%Vents vary greatly in length, with average lengths of $32$ tokens (words) when separating by space or punctuation (median$=17$, stdv=76, longest comment=18k tokens). We limit comment length to span between 3 and 32 tokens, and reduce the label space by filtering emotions that are marked as disabled or whose meaning is ambiguous, e.g. those containing emoji like %``{\NotoEmoji \symbol{"1F344}} Mushy {\NotoEmoji \symbol{"1F344}}''. 

\subsection{Multi-Label Benchmark Design}
\label{sec:BenchmarkingTextClass}

\begin{figure}[!ht]
    \centering
    \includegraphics[width=0.75\linewidth]{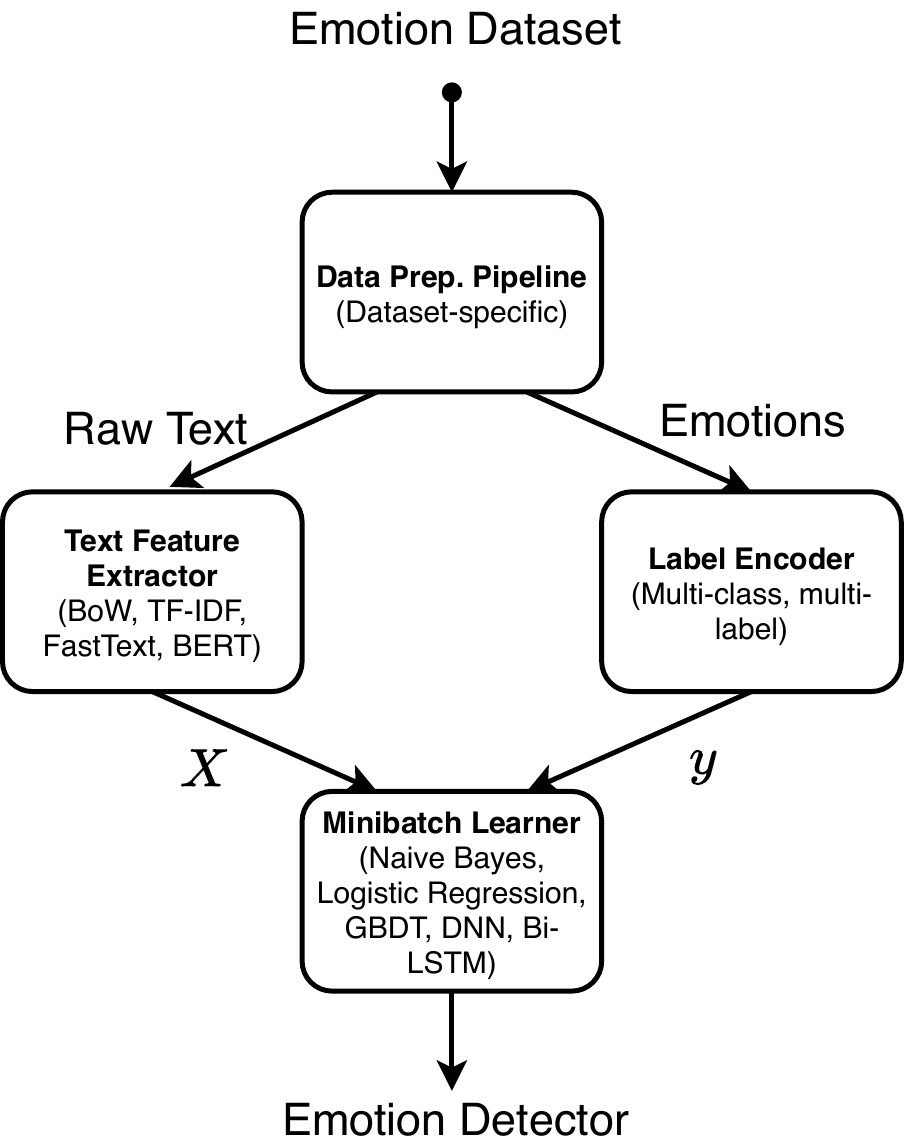}
    \caption{High level architecture. We implement several modules to transparently execute the different combinations of text representation and learning algorithms. The options between parenthesis detail the alternative implementations available on every step of the benchmark.}
    \label{fig:MethodologyArchitecture}
\end{figure}

We design a common architecture that lets us train, evaluate, and optionally transfer our models, and only use methods that may be applied in a streaming mini-batch fashion. Our approach is composed of three components, shown as sequential steps in \autoref{fig:MethodologyArchitecture}.
 
%\begin{asparaenum}
%    \item \textbf{dataset preparation.} \AKn{This point is a bit confusing an written with to many cross references. Maybe we do not really need that in the main paper } We normalize and unify the data to conform to a single interface. Our unification strategy maps text and labels to $\textbf{s}$ and $y$ from \autoref{sec:TaskDefinition} respectively. In this work, GoEmotions is already normalized while for Vent we follow the normalization procedure described in \autoref{sec:VentDataset}.
%    \item \textbf{Representation.} We transform the text and labels into vector representations that can be used in downstream machine learning tasks.
%    \item \textbf{Modelling.} Using the text and label representations, we train a predictive model to estimate the likelihood of the labels given an input text.
%\end{asparaenum}

%\AKn{I would put the following under the corresponding sections in the previous list.}
We implement two statistical and two embedding representation approaches, combined with five different learning algorithms. For statistical representations we use \textbf{Bag-of-Words} and \textbf{TF-IDF}, training \textbf{Naive Bayes}, \textbf{Logistic Regression}, and \textbf{Incremental Random Forests} models. For neural-LM representations, we only evaluate pre-trained English models, exploring word-level representations with \textbf{FastText} and contextualized representations using \textbf{BERT} with fine-tuning. 

\begin{figure}[!ht]
\centering
  \begin{subfigure}[t]{0.9\columnwidth}
  	\centering
    \includegraphics[width=\textwidth]{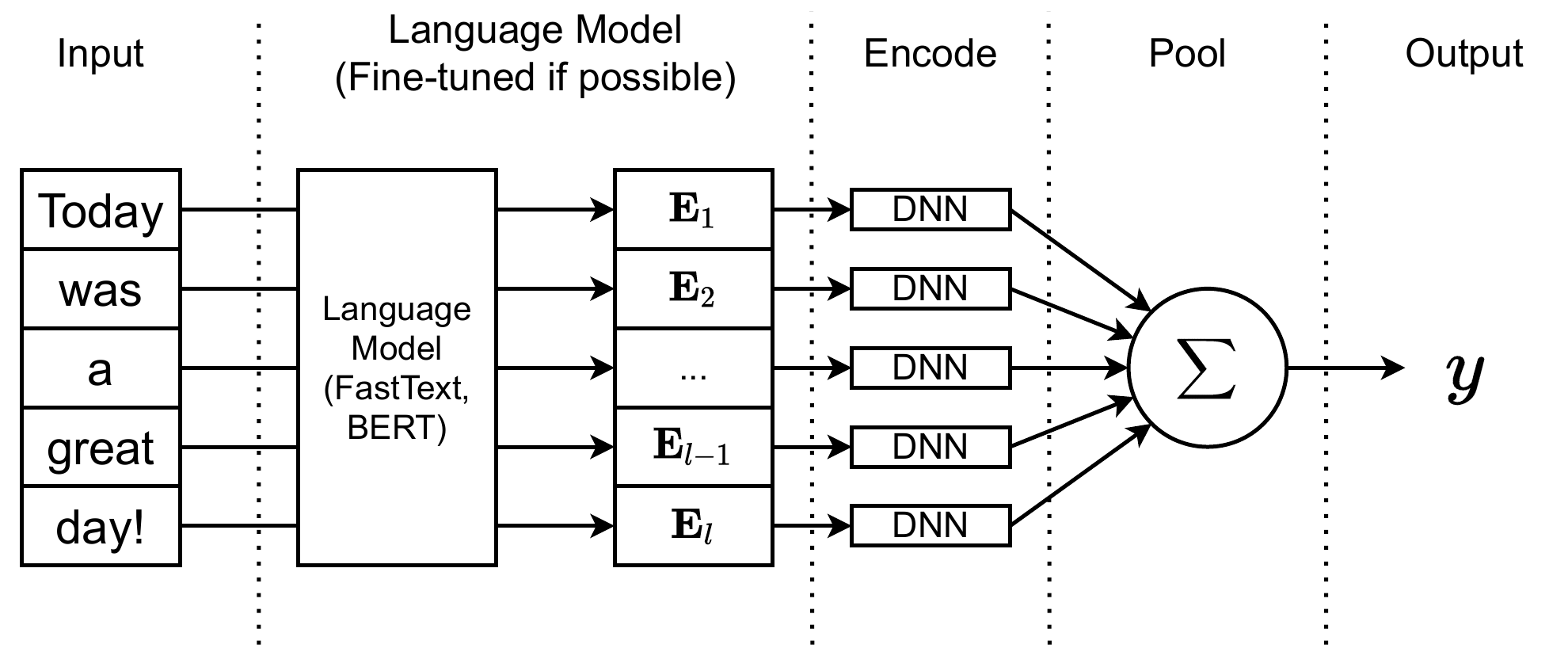}
    \caption{Pooled DNN Architecture.}
    \label{fig:DNNPoolArchitecture}
  \end{subfigure}
  \begin{subfigure}[t]{0.9\columnwidth}
  	\centering
    \includegraphics[width=\textwidth]{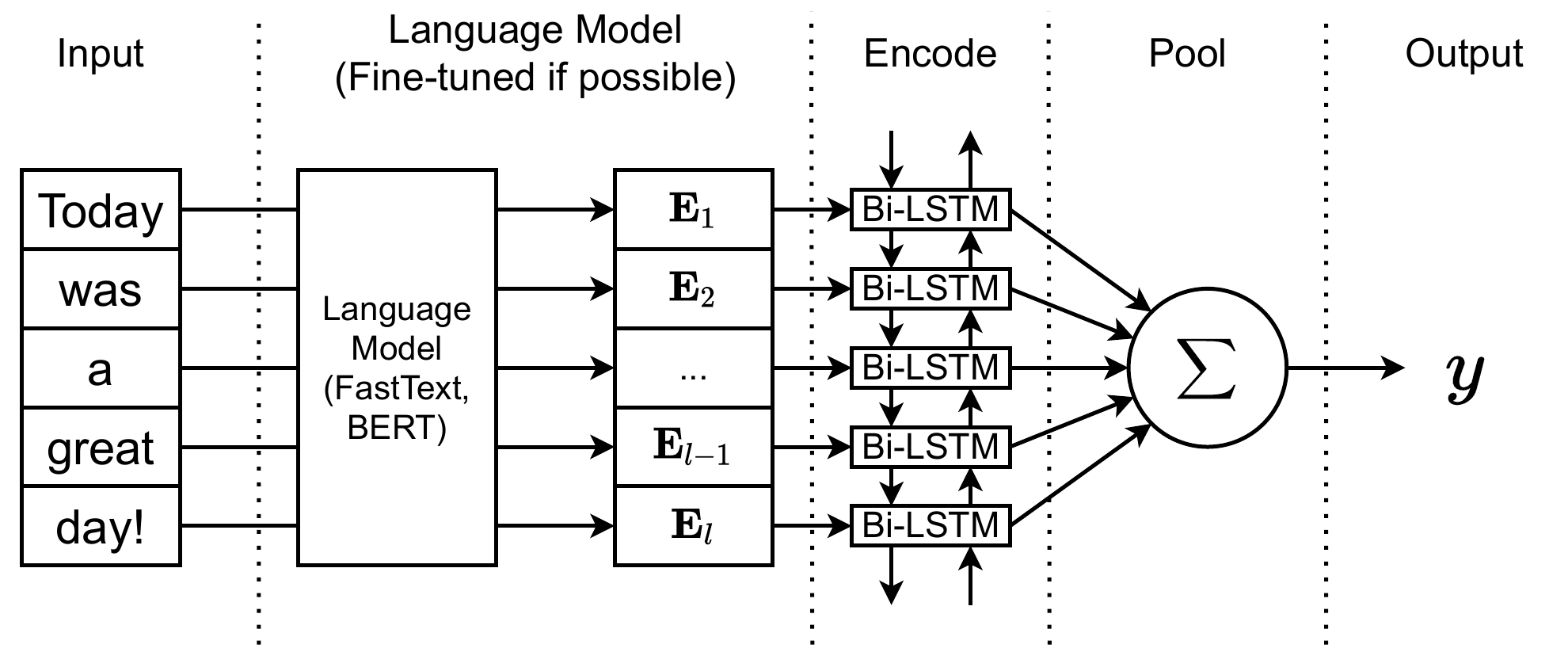}
    \caption{Bi-LSTM Architecture.}
    \label{fig:BiLSTMPoolArchitecture}
  \end{subfigure}
  \caption{Showcase of the neural architectures in the benchmark. BERT is fine-tuned while FastText is not due to implementation constraints.}
  \label{fig:NeuralArchitectures}
\end{figure}

As models for neural-LM representations, we design two simple neural architectures that receive embedding sequences as input and pool over output units to produce their outputs, as shown in~\autoref{fig:NeuralArchitectures}. We implement a multi-label classification objective, minimising the average binary cross-entropy loss over $N$ target emotions.

\begin{asparaenum}
    \item \textbf{(Pooled) Deep Neural Network.} We apply a DNN over every embedded token independently in parallel. 
    \item \textbf{(Bi)-LSTM.} We apply stacked (Bi)-LSTMs consuming the embedded sequence in a sequence to sequence manner. 
\end{asparaenum}

\section{Experimental Results}
\label{sec:Results}

%In this section we describe the results of our experiments and relate them to our research questions. 
In this section, we study the results of our benchmark, evaluate the behaviour of the trained models, and analyze the differences between \emph{writer} and \emph{reader} emotions on Vent.

\subsection{Emotion Benchmark Evaluation}

%\textbf{RQ1 - Modelling:}
First, we evaluate the emotion detection methods that we introduced in \autoref{sec:BenchmarkingTextClass}. We focus on four multi-label classification metrics: \textbf{M}acro F1, \textbf{m}icro F1, and micro-averaged \textbf{Pre}cision and \textbf{Rec}all. We use the original splits from the GoEmotion datasets, while we split Vent in 80 / 10 / 10 splits ordered by publication time for training, validation and testing. Our analysis is described on a per-dataset basis, with \autoref{tab:GoEmotionsBenchmark} and \autoref{tab:VentBenchmark} containing the results for GoEmotions and Vent respectively, showing  the best performing models in \textbf{bold}.

\begin{table}[!ht]
\centering
\begin{tabular}{@{}l@{\hskip 3pt}l@{}r@{\hskip 3pt}r@{\hskip 3pt}r@{\hskip 3pt}r@{}}
\toprule
\multicolumn{1}{@{}c}{\textbf{Repr.}} & \multicolumn{1}{c}{\textbf{Model}} & \multicolumn{1}{@{}c@{}}{\textbf{M-F1}} & \multicolumn{1}{c@{}}{\textbf{m-F1}} & \multicolumn{1}{c}{\textbf{Pre}} & \multicolumn{1}{c@{}}{\textbf{Rec}} \\ \midrule
\multirow{3}{*}{BoW}          & N. Bayes                        & 0.34                                 & 0.46 & 0.43 & 0.52                                \\
                                       & Log. Reg.                          & 0.45                                 & 0.53 & 0.48 & 0.61                                \\
                                       & R. Forest                      & 0.45                                 & 0.52 & 0.50 & 0.59                                \\\midrule
\multirow{3}{*}{TF-IDF}                & N. Bayes                        & 0.33                                 & 0.44 & 0.43 & 0.49                                \\
                                       & Log. Reg.                          & 0.47                                 & 0.53 & 0.49 & 0.60                                \\ 
                                       & R. Forest                      & 0.45                                 & 0.52 & 0.48 & 0.60                                \\\midrule
\multirow{2}{*}{FT} & DNN Pool & 0.42     & 0.49 & 0.45 & 0.61  \\
                          & Bi-LSTM  & 0.44     & 0.54 & 0.51 & 0.58  \\ \midrule
\multirow{3}{*}{BERT}     & Baseline\textsuperscript{*} & 0.46     & 0.51 & --- & ---  \\
                          & DNN Pool & \textbf{0.48}     & 0.55 & 0.52 & 0.61  \\
                          & Bi-LSTM  & 0.47     & \textbf{0.57} & \textbf{0.53} & \textbf{0.62}  \\ \bottomrule
\end{tabular}
{\scriptsize \textsuperscript{} \ \textsuperscript{*}Results as reported by~\cite{demszky2020}}
\caption{GoEmotions results. All stddevs $\leq$ 0.01.}
\label{tab:GoEmotionsBenchmark}
\end{table}

\begin{table}[!ht]
\centering
\begin{tabular}{@{}l@{\hskip 3pt}l@{}r@{\hskip 3pt}r@{\hskip 3pt}r@{\hskip 3pt}r@{}}
\toprule
\multicolumn{1}{@{}c}{\textbf{Repr.}} & \multicolumn{1}{c}{\textbf{Model}} & \multicolumn{1}{@{}c@{}}{\textbf{M-F1}} & \multicolumn{1}{c@{}}{\textbf{m-F1}} & \multicolumn{1}{c}{\textbf{Pre}} & \multicolumn{1}{c@{}}{\textbf{Rec}}
\\ \midrule
---                           & Random                 & 0.02                                 & 0.04 & 0.02 & 0.69 \\
\midrule
\multirow{3}{*}{BoW}          & N. Bayes                        & 0.13                                 & 0.15 & 0.12 & 0.23                                \\
                                       & Log. Reg.                          & 0.13                                 & 0.15 & 0.12 & 0.20                                \\
                                       & R. Forest                      & 0.11                                 & 0.13 & 0.12 & 0.18                                \\\midrule
\multirow{3}{*}{TF-IDF}                & N. Bayes                        & 0.14                                 & 0.16 & 0.13 & 0.21                                \\
                                       & Log. Reg.                          & 0.14                                 & 0.16 & 0.14 & 0.21                                \\ 
                                       & R. Forest                      & 0.11                                 & 0.13 & 0.12 & 0.19                                \\\midrule
\multirow{2}{*}{FT} & DNN Pool & 0.08     & 0.10 & 0.08 & 0.30  \\
                          & Bi-LSTM  & 0.08     & 0.11 & 0.08 & 0.21  \\ \midrule
\multirow{2}{*}{BERT}     & DNN Pool & 0.17     & 0.19 & 0.16 & 0.24  \\
                          & Bi-LSTM  & \textbf{0.19} & \textbf{0.21} & \textbf{0.19} & \textbf{0.26} \\ \bottomrule
\end{tabular}
\caption{Vent results. All stddevs $\leq$ 0.01, except FastText models. Random is a baseline classifier that produces a random score between 0 and 1 for every snippet and label.}
\label{tab:VentBenchmark}
\end{table}

 %\textbf{RQ1a - Model Benchmark:}
 %For both GoEmotions and Vent,
 BERT representations outperform every other configuration in both datasets. Furthermore, our results show that Logistic Regressions outperform Naive Bayes and Incremental Random Forests on configurations using Bag-of-Words or TF-IDF. On GoEmotions, our highest micro-F1 score is 0.57, while on Vent the highest micro-F1 score is 0.21. 

Our results on GoEmotions outperform the previous strong baseline using BERT, increasing macro-F1 from 0.46 to 0.47 and micro-F1 from 0.51 to 0.57. The positive results indicate that our design approach for the benchmark was appropriate to achieve a robust comparison between neural and non-neural methods. 

A possible explanation for the performance gains in our models over the baseline~\cite{demszky2020} is that our approach pools over each token in the input while the baseline uses the contextualized embedding of the last token in the sentence for its prediction. We observe a modest improvement from statistical methods over the baseline, which might be due to emotional information being largely encoded at the word level. 

On Vent, we find that the BERT model with pooled Bi-LSTM layers significantly outperform other methods. In comparison with GoEmotions, non-BERT models perform significantly worse than BERT-based models. A possible explanation for this is the noisy nature of venter-annotated text from social media posts, whose chosen emotions might at times be arbitrary, and the increased ambiguity from the 88 emotion labels in Vent compared to the 28 targets in GoEmotions. When we vary the fraction of the whole training set used for training, we find that statistical models perform closer to BERT when there is a reduced amount of data. A possible explanation for this behaviour might be due to our design approach, namely, by the introduction of layers on top of BERT that slow down convergence, as shown in~\autoref{fig:NeuralArchitectures}. 

To our surprise, we find that FastText consistently underperforms in comparison to statistical models. In particular, the TF-IDF logistic regression model significantly outperforms FastText on Vent. Furthermore, TF-IDF models using logistic regressions or random forests show similar performance to the FastText model using pooled Bi-LSTMs on GoEmotions despite the relative simplicity of the models. A possible explanation for the the observed difference in performance between Vent and GoEmotions might be that pretrained FastText models are unable to represent slang, typos and platform-specific vocabulary from Vent despite their underlying $n$-gram model.

%\subsubsection{Temporality in Vent} \NAn{Does this add anything valuable? Should we simply include this analysis in the sup. material?}

%Emotional data might not be stationary, e.g., the expressions used to convey emotion change over time and systems that presume a constant way of conveying emotion may degrade over time. We expect that the model trained over random splits will perform better than the one trained on temporal splits, as measured by F1-Score. This hypothesis will help practitioners understand whether or not emotion recognition models require frequent data collection and re-training strategies to stay relevant.

%\subsubsection{Emotional Transfer Learning}

%Given the size of the Vent dataset, we hypothesized training on it might  specialise BERT for emotion recognition tasks. If so, the a GoEmotions model trained on the fine-tuned BERT will perform better (as measured by micro F1-Score) than vanilla BERT trained directly on GoEmotions. We evaluate whether or not the hypothesis holds by implementing a transfer learning task from Vent to GoEmotions. Using the best performing BERT / Bi-LSTM configuration on different subsets of Vent, we take the finetuned BERT models and use them as embedding layers to repeat our experiments GoEmotions while changing no other hyper-parameter. However, we do not find any improvements on the transfer learning task, leading us to believe that the bulk of the signal might be encoded in the task-specific Bi-LSTM model rather than within the BERT layers. 

%\subsection{The Hierarchical Landscape of Emotions in Vent}
\subsection{Hierarchy of Emotions in Vent}
The Vent dataset includes a single emotion provided by the writer, in contrast to other datasets, e.g., GoEmotions, that provide readers annotations. However, our multi-label approach produces probabilities for each emotion given a text message. This allows us to analyze the complex structure of emotions represented by our model from Vent.

\begin{figure}[!ht]
    \centering
    \includegraphics[width=1.0\linewidth]{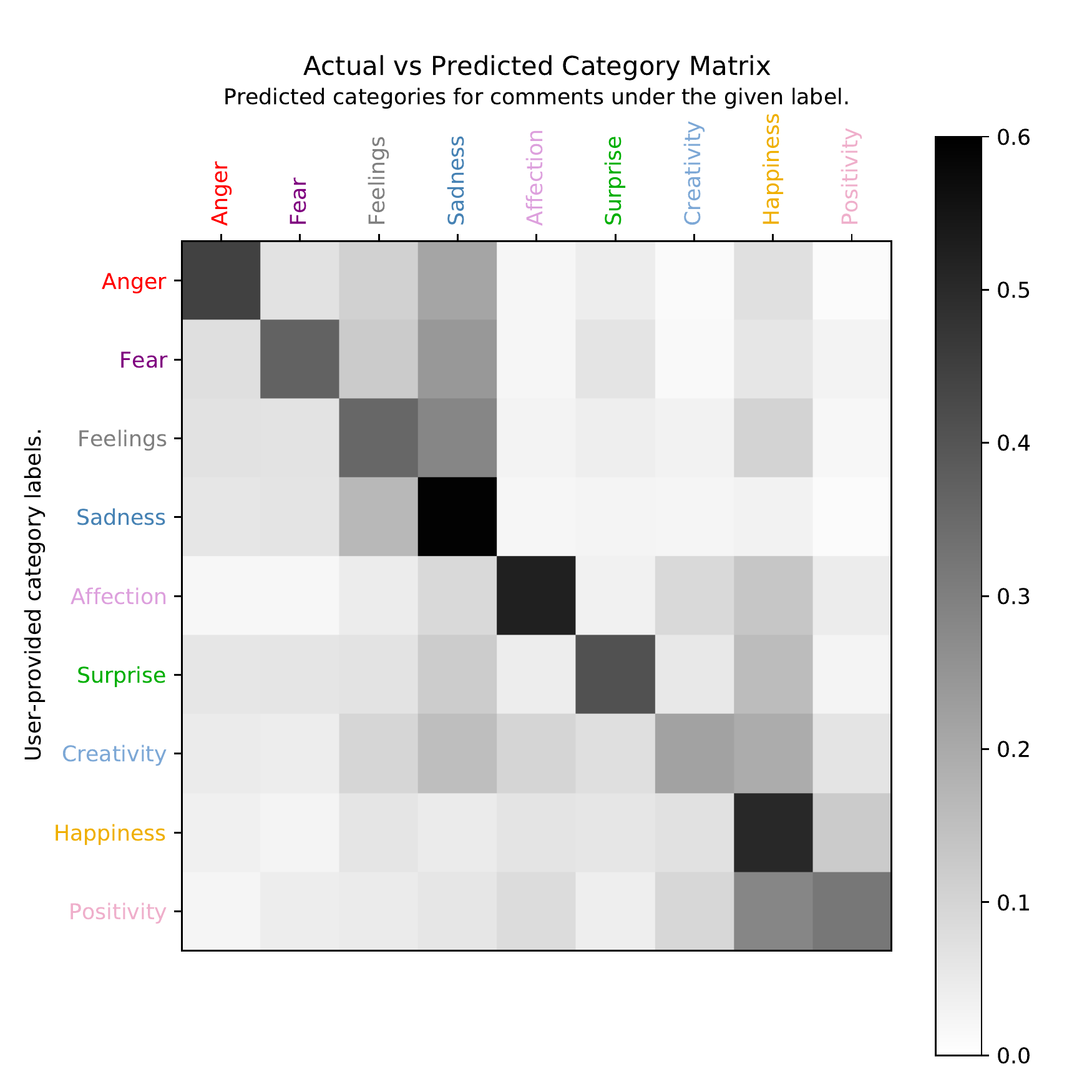}
    \caption{Normalized confusion matrix between categories, performing predictions by pooling over emotions as indicators of the category.}
    \label{fig:VentEmotionCategoryConfusionMatrix}
\end{figure}

\begin{figure*}[!t]
    \centering
    \begin{subfigure}[t]{1.0\textwidth}
        \centering
        \includegraphics[width=\textwidth]{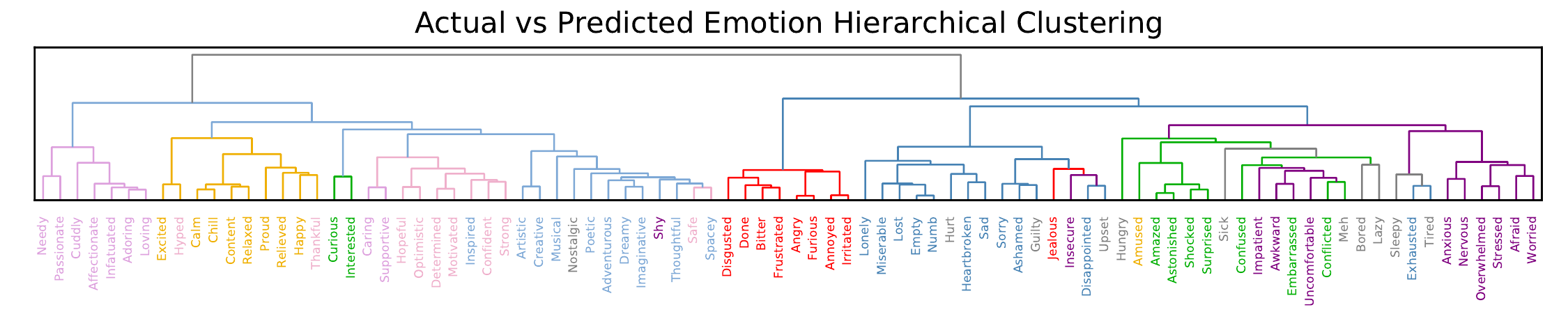} % Use Cropped if space is needed
        \label{fig:VentEmotionLabelDendrogram}
    \end{subfigure}
    \\[-2ex] % Negative space
    \begin{subfigure}[t]{1.0\textwidth}
        \centering
        \includegraphics[width=\textwidth]{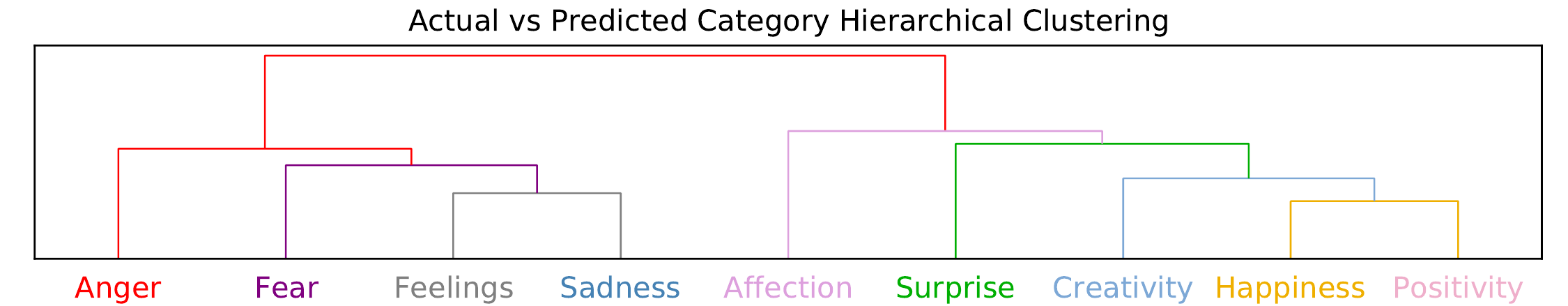} % Use Cropped if space is needed
        \label{fig:VentEmotionCategoryDendrogram}
    \end{subfigure}
    \\[-2ex] % Negative space
    \caption{Emotion (top) and Category (bottom) dendrograms obtained from the normalized activation for the best performing model captured by the emotion-level equivalent of~\autoref{fig:VentEmotionCategoryConfusionMatrix}. Color indicates emotion category.
    }
    \label{fig:VentEmotionDendrogram}
\end{figure*}

Instead of analysing pairwise correlations between emotions as is typically done, we build a (normalized) confusion matrix~$M$, where $M_{i,j}$ is the proportion of observing predicted emotion $j$ when a vent message is labelled with the $i$-th emotion. The resulting confusion matrix at the category level is shown in~\autoref{fig:VentEmotionCategoryConfusionMatrix} (see Supplementary Materials for the confusion matrix at the emotion level). We observe that negative categories (\textcolor[HTML]{FF0000}{Anger}, \textcolor[HTML]{800180}{Fear}, \textcolor[HTML]{808080}{Feelings}, and \textcolor[HTML]{4682B4}{Sadness}) and positive categories (\textcolor[HTML]{DDA0DD}{Affection}, \textcolor[HTML]{00AF02}{Surprise}, \textcolor[HTML]{7DA8D6}{Creativity}, \textcolor[HTML]{EFAF00}{Happiness}, and \textcolor[HTML]{EFAFCB}{Positivity}) are more often `confused' within each group than they are across both groups.

%We also study the predictive behaviour of the BERT / Bi-LSTM model. Interpretability efforts in deep models often focus on understanding the learned representations, and the biases, patterns, and semantics encoded within the latent spaces. In our case, we are concerned how the model predicts emotions, seeking to understand whether its behaviour resembles human confusion patterns or instead howcases surprising behaviour.

Each row in matrix $M$ represents an emotion activation pattern in the target space. We perform agglomerative clustering to discover the hierarchical structure of such space. \autoref{fig:VentEmotionDendrogram} shows the obtained dendrogram using the Euclidean distance between each pair of rows in $M$ as linkage metric. We show dendrograms at the emotion and category levels, max-pooling the activation patterns across emotion categories.

We observe that clusters of emotions generally align with emotion categories, despite category labels not being used during training. Top levels of the hierarchy clearly split positive and negative categories, while subsequent levels provide a sensible hierarchy of emotion clusters at different scales. At the emotion level, we find cases where clusters contain emotions belonging to different categories, but which appear to be semantically meaningful, e.g., the clusters formed by \textcolor[HTML]{FF0000}{Jealous}, \textcolor[HTML]{800180}{Insecure}, \textcolor[HTML]{4682B4}{Disappointed}, and \textcolor[HTML]{808080}{Upset}, or the cluster formed by \textcolor[HTML]{808080}{Sleepy}, \textcolor[HTML]{4682B4}{Exhausted}, and \textcolor[HTML]{808080}{Tired} in~\autoref{fig:VentEmotionDendrogram}.

\subsection{Experimental Evaluation}
We design a HIT task using Amazon's Mechanical Turk (MTurk) to evaluate the model and the differences between reader and writer emotions.
%~\cite{mturk}

\subsubsection{HIT Design}

We sample 30 Vent comments at random from the test split for each of the 88 emotions, building a dataset of 2,640 comments. We normalize snippets in the same manner as described in~\autoref{sec:ExperimentalDesign}. Additionally, we exclude comments that contain either \texttt{vent} or \texttt{nsfw}  (Not Safe For Work) terms to reduce the amount of self-referential or inappropriate content readers are exposed to. Comments are grouped in HITs of 10 comments, with 5 different readers assigned to each HIT, subject to an approval process to ensure quality annotations. We submit 264 HITs and receive work from 84 different readers with an average emotion accuracy of 11.43\% and an average category accuracy of 34.26\%. 

\Addition{Based on reader annotations, we analyse inter-annotator agreement and find that on average 1.81 ($\pm$ 0.88) readers out of the assigned 5 agree with the most frequently assigned emotion. At the category level, we find that 2.95 ($\pm$ 1.03) readers agree on average with the most frequent label. We find that the majority of readers can agree upon an emotion category, while generally not agreeing upon specific emotions within a category. We provide additional details on the inter-annotator agreement of readers in Section~\ref{app:MTurkAgreement}.}

The model outperforms human annotators (readers) on the subset of snippets submitted for labelling at the category level--0.395 vs 0.383 micro-F1, respectively. Readers show higher recall (0.725 vs 0.472) but lower precision (0.261 vs 0.356). This is expected, as we do not perform additional post-processing, i.e., filtering to select the majority label, to the readers' annotations.

The Vent dataset is collected from a social network with unknown user demographics, so emotion labels provided by writers might be noisy and/or biased. Therefore, we use reader annotations to study the appropriateness of Vent for learning emotions at scale. 

To do so, we invert the task: we predict the \emph{annotations provided by readers}, using the model trained on Vent. We seek to measure the overlap between the model, readers, and writers, evaluating whether the model aligns with what readers expect more than the original Vent labels (i.e. writers).

\begin{table}[!t]
\centering
\begin{tabular}{@{}c@{}c@{\hskip 4pt}c@{\hskip 3pt}r@{\hskip 3pt}r@{}}
\toprule
\textbf{Task}             & \textbf{Label} & \textbf{Predictor} & \multicolumn{1}{c@{}}{\textbf{M-F1}} & \multicolumn{1}{c@{}}{\textbf{m-F1}} \\ \midrule
\multirow{4}{*}{\textbf{Emotion}}& \multirow{2}{*}{\textbf{Writer}}                                                    & Reader         & 0.151                                  & 0.151                                  \\ 
                          &                & Model              & 0.181                                  & 0.181                                  \\ \cline{2-5} 
 
                          & \multirow{2}{*}{\textbf{Reader}}                                 & Model              & 0.241                                  & 0.246                                  \\ 
                          &                & Writer              & 0.151                                  & 0.136                                  \\ \midrule
\multirow{4}{*}{\textbf{Category}} & \multirow{2}{*}{\textbf{Writer}}                      & Reader         & 0.383                                  & 0.383                                  \\
                                   &              & Model              & 0.395                                  & 0.395                                  \\ \cline{2-5} 
                          & \multirow{2}{*}{\textbf{Reader}}                      & Model              & 0.467                                  & 0.471                                  \\ 
                          &                          & Writer              & 0.383                                  & 0.382                                  \\ \bottomrule
\end{tabular}
\caption{Comparison of results for predicting reader or writer labels ($N = 2640$) using our proposed model or the other available perspective.}
\label{tab:HITResults}
\end{table}

\autoref{tab:HITResults} shows emotion and category level macro- and micro-F1 scores for the model, readers, and writers. The model outperforms readers on the \emph{writer} prediction task--that is, when the labels are given by \emph{writers} and the predictors are either our model or readers--by 19.9\% and 3.1\% at the emotion and category levels in terms of relative micro-F1 (0.181 vs. 0.151 and 0.395 vs. 0.383 respectively). Surprisingly, the model is also capable of predicting the ambiguous emotional labels from \emph{readers} for the same texts, despite not being explicitly trained on the task. Specifically, the model achieves 35.9\% higher micro-F1 when evaluated against readers rather than writers at the emotion level (0.246 vs. 0.181), and 19.2\% at the category level (0.471 vs. 0.395). As a control experiment, we also compare with writer-provided emotions as a predictor of worker-provided emotions. We observe a micro-F1 gap which may be caused by non-uniform biases from readers towards certain emotions, which impact per-emotion support\footnote{Differences between readers and writers depend on the support (\# examples per target label) when micro-averaging, as values for precision/recall swap at the label level.}.

Our findings show that models trained on large amounts of writer-provided emotional labels from Vent are capable of capturing emotions perceived by readers. Our model achieves better performance when measured against readers rather than writers in terms of micro-F1, which aligns with existing literature on perspective-based emotion detection. For instance, previous work~\cite{buechel2017b} found readers' perspectives to be superior than writers' in terms of inter-annotator agreement on the VAD emotion model. These findings may explain the predictable nature of worker annotations, and the lower performances we observe when evaluating on writer-provided emotions. Our results are the first to consider both author and reader perspectives on categorical taxonomies \Addition{in large English corpora collected from social media}, which to our knowledge had not been studied in the emotion detection literature.

\section{Conclusions}
\label{sec:Conclusions}

We present a principled analysis of emotion detection techniques on the two largest available datasets to date: GoEmotions and Vent. Our thorough benchmark shows how different models behave, with BERT-based architectures consistently achieving the best performance across datasets. We release EmotionUI, a web interface for researchers to explore our models\footnote{
\href{http://emotionui.nur.systems/}{http://emotionui.nur.systems/}}\footnote{
\href{https://github.com/nur-ag/emotion-ui/}{https://github.com/nur-ag/emotion-ui/}}, and share our code, tools, annotations and experiment data\footnote{
\href{https://github.com/nur-ag/emotion-classification/}{https://github.com/nur-ag/emotion-classification/}}. 

On GoEmotions, our best performing model shows an improvement over the previous baseline by 11.8\% relative micro-F1. On Vent, we train a model that outperforms readers at predicting emotions provided by writers by 19.9\% relative micro-F1. Surprisingly, the same model shows better performance when evaluated against emotional labels provided by readers by 35.9\% relative micro-F1 despite not being trained on this task. Our findings show that models trained on Vent outperform readers in the task of predicting writer-provided emotions, even though the model performs better on reader-provided emotions rather than the original ground truth labels provided by Vent users (writers). These findings open new research directions on the automatic detection of emotions between readers and writers in discrete emotion taxonomies, and its implications for emotion detection systems. Our work suggests that the task of predicting the emotions that writers aim to express is harder than detecting those perceived by readers. However, the majority of current annotated emotion detection studies focus on reader emotions. 

\section{Acknowledgements}
\label{sec:Acknowledgements}

\Addition{
The project leading to these results has also received funding from ``La Caixa'' Foundation (ID 100010434), under the agreement LCF/PR/PR16 /51110009. 
Nurudin's research is partially funded by project 2018DI047 of the Catalan Industrial Doctorates Plan.
Vicenç G\'omez is supported by the Ramon y Cajal program RYC-2015-18878 (AEI/MINEICO/FSE,UE).
Andreas Kaltenbrunner acknowledges support from Intesa Sanpaolo Innovation Center. The funder had no role in study design, data collection and analysis, decision to publish, or preparation of the manuscript.
}

\section{Data Biases and Ethical Considerations}
\label{sec:EthicalConsiderations}

We present mechanisms to extract emotion information using text data from two social networks: Reddit and Vent. We believe that the user-generated nature of the data limits the generality of our results. For instance, it is understood that young adult males are over represented in the user base of Reddit~\cite{pew2016}. For Vent, no demographics data is available, which should further warrant caution on the part of researchers and practitioners building upon our work. Although there is evidence for language use differences across genders, age-groups and personality traits~\cite{schwartz2013}, data mined from social settings might amplify spurious relationships and lead to incomplete and biased accounts of such differences.

As emotional aspects of language are neither objective nor universal, we expect annotation biases in the target labels. In the case of GoEmotions, each Reddit comment was annotated by either 3 or 5 human judges given a list of 27 emotion definitions. However, the judges all share the same location, despite known cultural differences in the expression and understanding of emotions~\cite{scollon2004,jackson2019}. As such, the labels might not generalise across different cultures even when the language--English--is the same. In the case of Vent, emotions are provided by each person according to their own emotional state within a cultural group (the Vent community). In this sense, the usage patterns of may have shaped variation in how writers in Vent conceptualize emotions.

There are privacy and discrimination considerations on both data and models. The GoEmotions dataset is anonymised to ensure that the identity of the commenting users is kept safe. On Vent, user identifiers and hyperlinks in comments are masked to prevent user linking. However, due to the size of the dataset, it is possible that other personal details such as real names are contained in the complete text dumps. As such, the authors of \cite{lykousas2019} provide the dataset on a private at-request basis. Large language models retrained on either dataset, and particularly those trained on Vent, might end up encoding personal information in a way that can later be extracted~\cite{carlini2020}. However, we believe that in the context our work, the risk is reduced as we do not fine-tune the underlying language model in a generative task that would promote BERT to memorise parts of Vent.

A broader risk of our work is the potential emotion detection models to amplify and produce abuse. In recent years, generative methods to produce comments for news articles have been published at large academic venues~\cite{yang2019}, raising concerns of their potential to shape public perception or augment the reach of fake news stories. We believe that emotion detection systems may be used to further enhance comment generation models, allowing their designers to adversarially craft content aiming shape the emotional perceptions of readers. Models trained on a large-scale dataset such as Vent might be used to condition generated comments so that they foster a desired kind of emotional discourse, positive or negative. Because of this, we will provide our models on request to interested researchers rather than making them fully available upon publication to limit their usage by unknown third parties.

\bibliography{references}

\begin{thebibliography}{53}
\expandafter\ifx\csname natexlab\endcsname\relax\def\natexlab#1{#1}\fi

\bibitem[{Abdul-Mageed and Ungar(2017)}]{abdul2017}
Muhammad Abdul-Mageed and Lyle Ungar. 2017.
\newblock {E}mo{N}et: Fine-grained emotion detection with gated recurrent
  neural networks.
\newblock In \emph{55th Annual Meeting of the Association for Computational
  Linguistics}.

\bibitem[{Acheampong et~al.(2021)Acheampong, Nunoo-Mensah, and
  Chen}]{acheampong2021}
Francisca~A Acheampong, Henry Nunoo-Mensah, and Wenyu Chen. 2021.
\newblock \href {https://doi.org/10.1007/s10462-021-09958-2} {Transformer
  models for text-based emotion detection: a review of bert-based approaches}.
\newblock \emph{Artificial Intelligence Review}, pages 1--41.

\bibitem[{Acheampong et~al.(2020)Acheampong, Wenyu, and
  Nunoo-Mensah}]{acheampong2020}
Francisca~Adoma Acheampong, Chen Wenyu, and Henry Nunoo-Mensah. 2020.
\newblock \href {https://doi.org/https://doi.org/10.1002/eng2.12189}
  {Text-based emotion detection: Advances, challenges, and opportunities}.
\newblock \emph{Engineering Reports}, 2(7):e12189.

\bibitem[{Agrawal and Awekar(2018)}]{agrawal2018}
Sweta Agrawal and Amit Awekar. 2018.
\newblock Deep learning for detecting cyberbullying across multiple social
  media platforms.
\newblock In \emph{Advances in Information Retrieval}, pages 141--153. Springer
  International Publishing.

\bibitem[{Alswaidan and Menai(2020)}]{alswaidan2020}
Nourah Alswaidan and Mohamed Menai. 2020.
\newblock \href {https://doi.org/10.1007/s10115-020-01449-0} {A survey of
  state-of-the-art approaches for emotion recognition in text}.
\newblock \emph{Knowledge and Information Systems}, 62.

\bibitem[{Bojanowski et~al.(2017)Bojanowski, Grave, Joulin, and
  Mikolov}]{bojanowski2017}
Piotr Bojanowski, Edouard Grave, Armand Joulin, and Tomas Mikolov. 2017.
\newblock \href {https://doi.org/10.1162/tacl_a_00051} {Enriching word vectors
  with subword information}.
\newblock \emph{Transactions of the Association for Computational Linguistics},
  5:135--146.

\bibitem[{Bostan and Klinger(2018)}]{bostan2018}
Laura-Ana-Maria Bostan and Roman Klinger. 2018.
\newblock An analysis of annotated corpora for emotion classification in text.
\newblock In \emph{27th International Conference on Computational Linguistics}.

\bibitem[{Brassard-Gourdeau and Khoury(2019)}]{brassard-gourdeau2019}
Eloi Brassard-Gourdeau and Richard Khoury. 2019.
\newblock Subversive toxicity detection using sentiment information.
\newblock In \emph{Third Workshop on Abusive Language Online}.

\bibitem[{Buechel and Hahn(2017{\natexlab{a}})}]{buechel2017b}
Sven Buechel and Udo Hahn. 2017{\natexlab{a}}.
\newblock {E}mo{B}ank: Studying the impact of annotation perspective and
  representation format on dimensional emotion analysis.
\newblock In \emph{15th Conference of the {E}uropean Chapter of the Association
  for Computational Linguistics}.

\bibitem[{Buechel and Hahn(2017{\natexlab{b}})}]{buechel2017}
Sven Buechel and Udo Hahn. 2017{\natexlab{b}}.
\newblock Readers vs. writers vs. texts: Coping with different perspectives of
  text understanding in emotion annotation.
\newblock In \emph{11th Linguistic Annotation Workshop}, pages 1--12.

\bibitem[{Carlini et~al.(2020)Carlini, Tram{\`{e}}r, Wallace, Jagielski,
  Herbert{-}Voss, Lee, Roberts, Brown, Song, Erlingsson, Oprea, and
  Raffel}]{carlini2020}
Nicholas Carlini, Florian Tram{\`{e}}r, Eric Wallace, Matthew Jagielski, Ariel
  Herbert{-}Voss, Katherine Lee, Adam Roberts, Tom~B. Brown, Dawn Song,
  {\'{U}}lfar Erlingsson, Alina Oprea, and Colin Raffel. 2020.
\newblock \href {http://arxiv.org/abs/2012.07805} {Extracting training data
  from large language models}.
\newblock \emph{CoRR}, abs/2012.07805.

\bibitem[{Center(2016)}]{pew2016}
Pew~Research Center. 2016.
\newblock Nearly eight-in-ten {Reddit} users get news on the site.

\bibitem[{Chauhan et~al.(2020)Chauhan, R, Ekbal, and
  Bhattacharyya}]{chauhan2020}
Dushyant~Singh Chauhan, Dhanush~S R, Asif Ekbal, and Pushpak Bhattacharyya.
  2020.
\newblock \href {https://doi.org/10.18653/v1/2020.acl-main.401} {Sentiment and
  emotion help sarcasm? {A} multi-task learning framework for multi-modal
  sarcasm, sentiment and emotion analysis}.
\newblock In \emph{Proceedings of the 58th Annual Meeting of the Association
  for Computational Linguistics}, pages 4351--4360, Online. Association for
  Computational Linguistics.

\bibitem[{Cowen et~al.(2018)Cowen, Elfenbein, Laukka, and Keltner}]{cowen2018}
Alan Cowen, Hillary Elfenbein, Petri Laukka, and Dacher Keltner. 2018.
\newblock \href {https://doi.org/10.1037/amp0000399} {Mapping 24 emotions
  conveyed by brief human vocalization}.
\newblock \emph{American Psychologist}.

\bibitem[{Cowen and Keltner(2019)}]{cowen2019b}
Alan Cowen and Dacher Keltner. 2019.
\newblock \href {https://doi.org/10.1037/amp0000488} {What the face displays:
  Mapping 28 emotions conveyed by naturalistic expression}.
\newblock \emph{American Psychologist}, 75.

\bibitem[{Cowen et~al.(2019)Cowen, Sauter, Tracy, and Keltner}]{cowen2019}
Alan Cowen, Disa Sauter, Jessica~L. Tracy, and Dacher Keltner. 2019.
\newblock Mapping the passions: Toward a high-dimensional taxonomy of emotional
  experience and expression.
\newblock \emph{Psychological Science in the Public Interest}, 20(1):69--90.

\bibitem[{Cowen et~al.(2020)Cowen, Fang, Sauter, and Keltner}]{cowen2020}
Alan~S. Cowen, Xia Fang, Disa Sauter, and Dacher Keltner. 2020.
\newblock \href {https://doi.org/10.1073/pnas.1910704117} {What music makes us
  feel: At least 13 dimensions organize subjective experiences associated with
  music across different cultures}.
\newblock \emph{Proceedings of the National Academy of Sciences},
  117(4):1924--1934.

\bibitem[{Crowdflower(2016)}]{crowdflower2016}
Crowdflower. 2016.
\newblock The emotion in text.
\newblock
  \url{https://www.figure-eight.com/data/sentiment-analysis-emotion-text/}.

\bibitem[{Danescu-Niculescu-Mizil et~al.(2013)Danescu-Niculescu-Mizil, West,
  Jurafsky, Leskovec, and Potts}]{danescu2013}
Cristian Danescu-Niculescu-Mizil, Robert West, Dan Jurafsky, Jure Leskovec, and
  Christopher Potts. 2013.
\newblock \href {https://doi.org/10.1145/2488388.2488416} {No country for old
  members: User lifecycle and linguistic change in online communities}.
\newblock In \emph{Proceedings of the 22nd International Conference on World
  Wide Web}, WWW '13, page 307–318, New York, NY, USA. Association for
  Computing Machinery.

\bibitem[{Demszky et~al.(2020)Demszky, Movshovitz-Attias, Ko, Cowen, Nemade,
  and Ravi}]{demszky2020}
Dorottya Demszky, Dana Movshovitz-Attias, Jeongwoo Ko, Alan Cowen, Gaurav
  Nemade, and Sujith Ravi. 2020.
\newblock {G}o{E}motions: A dataset of fine-grained emotions.
\newblock In \emph{58th Annual Meeting of the Association for Computational
  Linguistics}.

\bibitem[{Devlin et~al.(2019)Devlin, Chang, Lee, and Toutanova}]{devlin2019}
Jacob Devlin, Ming-Wei Chang, Kenton Lee, and Kristina Toutanova. 2019.
\newblock {BERT}: Pre-training of deep bidirectional transformers for language
  understanding.
\newblock In \emph{Proceedings of the 2019 Conference of the North {A}merican
  Chapter of the Association for Computational Linguistics: Human Language
  Technologies}, pages 4171--4186.

\bibitem[{Ekman and Friesen(1971)}]{ekman1971}
Paul Ekman and Wallace~V. Friesen. 1971.
\newblock \href {https://doi.org/10.1037/h0030377} {Constants across cultures
  in the face and emotion.}
\newblock \emph{Journal of Personality and Social Psychology}, 17(2):124--129.

\bibitem[{{El Ayadi} et~al.(2011){El Ayadi}, Kamel, and Karray}]{elayadi2011}
Moataz {El Ayadi}, Mohamed~S. Kamel, and Fakhri Karray. 2011.
\newblock \href {https://doi.org/https://doi.org/10.1016/j.patcog.2010.09.020}
  {Survey on speech emotion recognition: Features, classification schemes, and
  databases}.
\newblock \emph{Pattern Recognition}, 44(3):572 -- 587.

\bibitem[{Fung et~al.(2018)Fung, Bertero, Wan, Dey, Chan, Bin~Siddique, Yang,
  Wu, and Lin}]{fung2016}
Pascale Fung, Dario Bertero, Yan Wan, Anik Dey, Ricky Ho~Yin Chan, Farhad
  Bin~Siddique, Yang Yang, Chien-Sheng Wu, and Ruixi Lin. 2018.
\newblock Towards empathetic human-robot interactions.
\newblock In \emph{Computational Linguistics and Intelligent Text Processing},
  pages 173--193.

\bibitem[{(Garethjns)()}]{incremental-trees}
Gareth~Jones (Garethjns).
\newblock \href {https://github.com/garethjns/IncrementalTrees}
  {Incremental{T}rees: {O}nline {T}ree {L}earners on top of {S}cikit-learn}.
\newblock \url{https://github.com/garethjns/IncrementalTrees}.

\bibitem[{Guerini and Staiano(2015)}]{guerini2015}
Marco Guerini and Jacopo Staiano. 2015.
\newblock Deep feelings: A massive cross-lingual study on the relation between
  emotions and virality.
\newblock In \emph{24th International Conference on World Wide Web}.

\bibitem[{Husseini~Orabi et~al.(2018)Husseini~Orabi, Buddhitha, Husseini~Orabi,
  and Inkpen}]{husseini2018}
Ahmed Husseini~Orabi, Prasadith Buddhitha, Mahmoud Husseini~Orabi, and Diana
  Inkpen. 2018.
\newblock Deep learning for depression detection of {T}witter users.
\newblock In \emph{Fifth Workshop on Computational Linguistics and Clinical
  Psychology: From Keyboard to Clinic}, pages 88--97.

\bibitem[{Hutto and Gilbert(2014)}]{hutto2014}
Clayton Hutto and Eric Gilbert. 2014.
\newblock Vader: A parsimonious rule-based model for sentiment analysis of
  social media text.
\newblock In \emph{7th International AAAI Conference on Web and Social Media},
  volume~8.

\bibitem[{Jackson et~al.(2019)Jackson, Watts, Henry, List, Forkel, Mucha,
  Greenhill, Gray, and Lindquist}]{jackson2019}
Joshua~Conrad Jackson, Joseph Watts, Teague~R. Henry, Johann-Mattis List,
  Robert Forkel, Peter~J. Mucha, Simon~J. Greenhill, Russell~D. Gray, and
  Kristen~A. Lindquist. 2019.
\newblock \href {https://doi.org/10.1126/science.aaw8160} {Emotion semantics
  show both cultural variation and universal structure}.
\newblock \emph{Science}, 366(6472):1517--1522.

\bibitem[{{Li} and {Deng}(2020)}]{li2020}
S.~{Li} and W.~{Deng}. 2020.
\newblock \href {https://doi.org/10.1109/TAFFC.2020.2981446} {Deep facial
  expression recognition: A survey}.
\newblock \emph{IEEE Transactions on Affective Computing}, pages 1--1.

\bibitem[{Lykousas et~al.(2019)Lykousas, Patsakis, Kaltenbrunner, and
  Gómez}]{lykousas2019}
Nikolaos Lykousas, Constantinos Patsakis, Andreas Kaltenbrunner, and Vicenç
  Gómez. 2019.
\newblock \href {https://www.aaai.org/ojs/index.php/ICWSM/article/view/3361}
  {Sharing emotions at scale: The vent dataset}.
\newblock \emph{13th International AAAI Conference on Weblogs and Social
  Media}, 13(01):611--619.

\bibitem[{Malko et~al.(2021)Malko, Paris, Duenser, Kangas, Molla, Sparks, and
  Wan}]{malko-etal-2021-demonstrating}
Anton Malko, Cecile Paris, Andreas Duenser, Maria Kangas, Diego Molla, Ross
  Sparks, and Stephen Wan. 2021.
\newblock \href {https://doi.org/10.18653/v1/2021.clpsych-1.5} {Demonstrating
  the reliability of self-annotated emotion data}.
\newblock In \emph{Proceedings of the Seventh Workshop on Computational
  Linguistics and Clinical Psychology: Improving Access}, pages 45--54, Online.
  Association for Computational Linguistics.

\bibitem[{Mohammad(2021)}]{mohammad2021sentiment}
Saif~M Mohammad. 2021.
\newblock Sentiment analysis: Automatically detecting valence, emotions, and
  other affectual states from text.
\newblock In \emph{Emotion Measurement}, pages 323--379. Elsevier.

\bibitem[{Mohammad et~al.(2013)Mohammad, Kiritchenko, and Zhu}]{mohammad2013}
Saif~M Mohammad, Svetlana Kiritchenko, and Xiaodan Zhu. 2013.
\newblock {NRC}-{C}anada: Building the state-of-the-art in sentiment analysis
  of tweets.
\newblock In \emph{Second Joint Conference on Lexical and Computational
  Semantics}, pages 321--327.

\bibitem[{Mohammad and Turney(2013)}]{mohammad2013b}
Saif~M Mohammad and Peter Turney. 2013.
\newblock Crowdsourcing a word-emotion association lexicon.
\newblock \emph{Computational Intelligence}, 29.

\bibitem[{Mohammad et~al.(2015)Mohammad, Zhu, Kiritchenko, and
  Martin}]{mohammad2015}
Saif~M. Mohammad, Xiaodan Zhu, Svetlana Kiritchenko, and Joel Martin. 2015.
\newblock Sentiment, emotion, purpose, and style in electoral tweets.
\newblock \emph{Information Processing and Management}, 51(4):480–499.

\bibitem[{Nielsen(2011)}]{nielsen2011}
Finn Nielsen. 2011.
\newblock A new {ANEW:} evaluation of a word list for sentiment analysis in
  microblogs.
\newblock In \emph{{ESWC2011} Workshop on 'Making Sense of Microposts'}, volume
  718 of \emph{{CEUR} Workshop Proceedings}, pages 93--98.

\bibitem[{Noroozi et~al.(2018)Noroozi, Kaminska, Corneanu, Sapinski, Escalera,
  and Anbarjafari}]{noroozi2018}
Fatemeh Noroozi, Dorota Kaminska, Ciprian Corneanu, Tomasz Sapinski, Sergio
  Escalera, and Gholamreza Anbarjafari. 2018.
\newblock Survey on emotional body gesture recognition.
\newblock \emph{IEEE transactions on affective computing}.

\bibitem[{Ortigosa et~al.(2014)Ortigosa, Martín, and Carro}]{ortigosa2014}
Alvaro Ortigosa, José~M. Martín, and Rosa~M. Carro. 2014.
\newblock \href {https://doi.org/https://doi.org/10.1016/j.chb.2013.05.024}
  {Sentiment analysis in facebook and its application to e-learning}.
\newblock \emph{Computers in Human Behavior}, 31:527 -- 541.

\bibitem[{Pedregosa et~al.(2011)Pedregosa, Varoquaux, Gramfort, Michel,
  Thirion, Grisel, Blondel, Prettenhofer, Weiss, Dubourg, Vanderplas, Passos,
  Cournapeau, Brucher, Perrot, and Duchesnay}]{scikit-learn}
F.~Pedregosa, G.~Varoquaux, A.~Gramfort, V.~Michel, B.~Thirion, O.~Grisel,
  M.~Blondel, P.~Prettenhofer, R.~Weiss, V.~Dubourg, J.~Vanderplas, A.~Passos,
  D.~Cournapeau, M.~Brucher, M.~Perrot, and E.~Duchesnay. 2011.
\newblock Scikit-learn: Machine learning in {P}ython.
\newblock \emph{Journal of Machine Learning Research}, 12:2825--2830.

\bibitem[{Plutchik(1980)}]{plutchik1980}
Robert Plutchik. 1980.
\newblock \href
  {https://doi.org/https://doi.org/10.1016/B978-0-12-558701-3.50007-7} {A
  general psychoevolutionary theory of emotion}.
\newblock In Robert Plutchik and Henry Kellerman, editors, \emph{Theories of
  Emotion}, pages 3 -- 33. Academic Press.

\bibitem[{Poria et~al.(2019)Poria, Majumderd, Mihalceae, and Hovy}]{poria2019}
Soujanya Poria, Navonil Majumderd, Rada Mihalceae, and Eduard Hovy. 2019.
\newblock \href {https://doi.org/10.1109/ACCESS.2019.2929050} {Emotion
  recognition in conversation: Research challenges, datasets, and recent
  advances}.
\newblock \emph{IEEE Access}, PP:1--1.

\bibitem[{Ptaszynski et~al.(2009)Ptaszynski, Dybala, Shi, Rzepka, and
  Araki}]{Ptaszynski2009}
M.~Ptaszynski, Pawel Dybala, Wenhan Shi, Rafal Rzepka, and K.~Araki. 2009.
\newblock A system for affect analysis of utterances in japanese supported with
  web mining.
\newblock \emph{Journal of Japan Society for Fuzzy Theory and Intelligent
  Informatics}, 21:194--213.

\bibitem[{Ram{\'i}rez-Cifuentes et~al.(2020)Ram{\'i}rez-Cifuentes, Freire,
  Baeza-Yates, Punt{\'i}, Medina-Bravo, Velazquez, Gonfaus, and
  Gonz{\`a}lez}]{jimir20}
Diana Ram{\'i}rez-Cifuentes, Ana Freire, Ricardo Baeza-Yates, Joaquim
  Punt{\'i}, Pilar Medina-Bravo, Diego~Alejandro Velazquez, Josep~Maria
  Gonfaus, and Jordi Gonz{\`a}lez. 2020.
\newblock Detection of suicidal ideation on social media: Multimodal,
  relational, and behavioral analysis.
\newblock \emph{Journal of Medical Internet Research}, 22(7).

\bibitem[{Russell(2003)}]{rusell2003}
James Russell. 2003.
\newblock \href {https://doi.org/10.1037//0033-295X.110.1.145} {Core affect and
  the psychological construction of emotion}.
\newblock \emph{Psychological review}, 110:145--72.

\bibitem[{Schwartz et~al.(2013)Schwartz, Eichstaedt, Kern, Dziurzynski,
  Ramones, Agrawal, Shah, Kosinski, Stillwell, Seligman, and
  Ungar}]{schwartz2013}
H.~Andrew Schwartz, Johannes~C. Eichstaedt, Margaret~L. Kern, Lukasz
  Dziurzynski, Stephanie~M. Ramones, Megha Agrawal, Achal Shah, Michal
  Kosinski, David Stillwell, Martin E.~P. Seligman, and Lyle~H. Ungar. 2013.
\newblock \href {https://doi.org/10.1371/journal.pone.0073791} {Personality,
  gender, and age in the language of social media: The open-vocabulary
  approach}.
\newblock \emph{PLOS ONE}, 8(9):1--16.

\bibitem[{Scollon et~al.(2004)Scollon, Diener, Oishi, and
  Biswas-Diener}]{scollon2004}
Christie~N. Scollon, Ed~Diener, Shigehiro Oishi, and Robert Biswas-Diener.
  2004.
\newblock \href {https://doi.org/10.1177/0022022104264124} {Emotions across
  cultures and methods}.
\newblock \emph{Journal of Cross-Cultural Psychology}, 35(3):304--326.

\bibitem[{Shmueli and Ku(2019)}]{shmueli2019}
Boaz Shmueli and Lun-Wei Ku. 2019.
\newblock Socialnlp emotionx 2019 challenge overview: Predicting emotions in
  spoken dialogues and chats.

\bibitem[{Silva et~al.(2014)Silva, Hruschka, and Hruschka}]{silva2014}
N.~F.~D. Silva, E.~Hruschka, and Estevam~R. Hruschka. 2014.
\newblock Tweet sentiment analysis with classifier ensembles.
\newblock \emph{Decis. Support Syst.}, 66:170--179.

\bibitem[{Strapparava and Mihalcea(2007)}]{strapparava2007}
Carlo Strapparava and Rada Mihalcea. 2007.
\newblock \href {https://www.aclweb.org/anthology/S07-1013} {{S}em{E}val-2007
  task 14: Affective text}.
\newblock In \emph{Proceedings of the Fourth International Workshop on Semantic
  Evaluations ({S}em{E}val-2007)}, pages 70--74, Prague, Czech Republic.
  Association for Computational Linguistics.

\bibitem[{Tausczik and Pennebaker(2010)}]{tausczik2010}
Yla Tausczik and James Pennebaker. 2010.
\newblock \href {https://doi.org/10.1177/0261927X09351676} {The psychological
  meaning of words: Liwc and computerized text analysis methods}.
\newblock \emph{Journal of Language and Social Psychology}, 29:24--54.

\bibitem[{Thelwall et~al.(2010)Thelwall, Buckley, Paltoglou, Cai, and
  Kappas}]{thelwall2010}
Mike Thelwall, Kevan Buckley, Georgios Paltoglou, Di~Cai, and Arvid Kappas.
  2010.
\newblock \href {https://doi.org/10.1002/asi.21416} {Sentiment strength
  detection in short informal text}.
\newblock \emph{Journal of the American Society for Information Science and
  Technology}, 61:2544--2558.

\bibitem[{Yang et~al.(2019)Yang, Xu, Wu, and Li}]{yang2019}
Ze~Yang, Can Xu, Wei Wu, and Zhoujun Li. 2019.
\newblock Read, attend and comment: A deep architecture for automatic news
  comment generation.
\newblock In \emph{Conference on Empirical Methods in Natural Language
  Processing and the 9th International Joint Conference on Natural Language
  Processing (EMNLP-IJCNLP)}.

\end{thebibliography}
\bibliographystyle{acl_natbib}

\onecolumn
\newpage
\twocolumn
\begin{appendices}

\section{Dataset Details}

\subsection{GoEmotions}

The GoEmotions dataset is built from comments sampled from a Reddit dump from 2005 to 2019. Snippets are filtered to (a) reduce profanity and offensive or discriminating content, (b) keep a consistent comment length of under 30 tokens with a median of 12 tokens per comment and (c) balance sentiment, emotion and subreddit popularities. Proper names and religions are masked with special, replacing their occurrences in the text with \texttt{[NAME]} and \texttt{[RELIGION]} respectively. 

\begin{table}[!hb]
\centering
\begin{tabular}{@{}ll@{}}
\toprule
\textbf{\# examples} & 58,009 \\
\textbf{\# emotions} & 27 + neutral \\
\textbf{\# unique raters} & 82 \\
\textbf{\# raters / example} & 3 or 5 \\ \hline
\textbf{\begin{tabular}[c]{@{}l@{}}Marked unclear\\ or difficult to label\end{tabular}} & 1.6\% \\ \hline
\multirow{4}{*}{\textbf{Labels per example}} & 1: 83\% \\
 & 2: 15\% \\
 & 3: 2\% \\
 & 4+: .2\% \\ \hline
\textbf{\begin{tabular}[c]{@{}l@{}}\# examples w/ 2+ raters\\ agreeing on at least 1 label\end{tabular}} & 54,263 (94\%) \\
\textbf{\begin{tabular}[c]{@{}l@{}}\# examples w/ 3+ raters\\ agreeing on at least 1 label\end{tabular}} & 17,763 (31\%) \\ \bottomrule
\end{tabular}
\caption{General dataset and annotation statistics for the GoEmotions dataset. Taken from~\cite{demszky2020}.}
\label{tab:GoEmotionStatistics}
\end{table}

\subsection{Vent}

\begin{figure}[!ht]
\centering
  \begin{subfigure}[t]{0.49\columnwidth}
  	\centering
    \includegraphics[width=\textwidth]{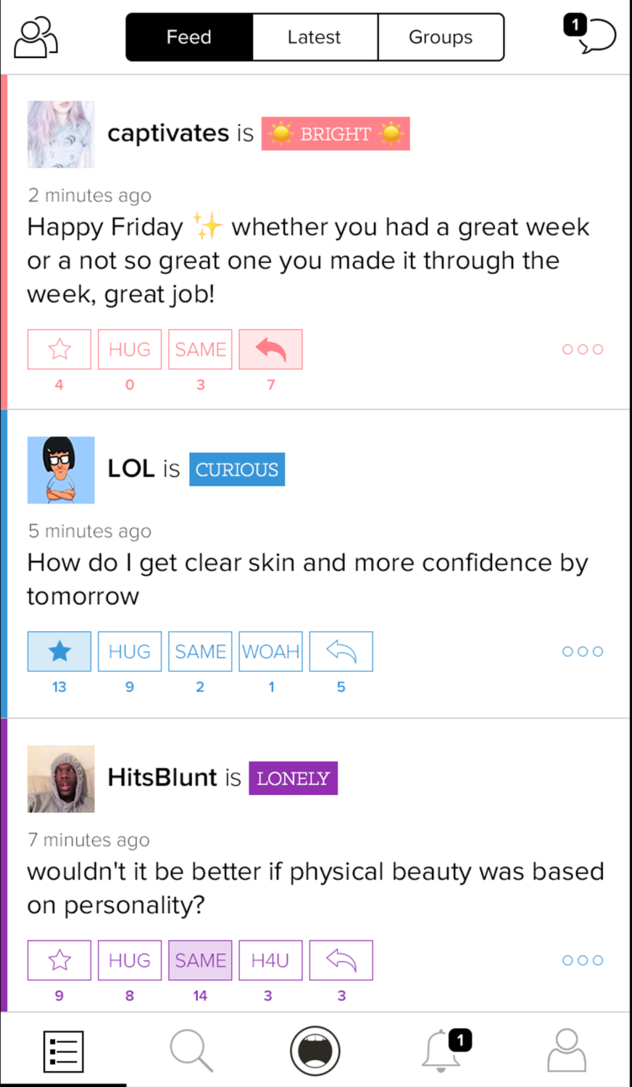}
    \caption{Vent Feed}
    \label{fig:VentFeed}
  \end{subfigure}
  \begin{subfigure}[t]{0.49\columnwidth}
  	\centering
    \includegraphics[width=\textwidth]{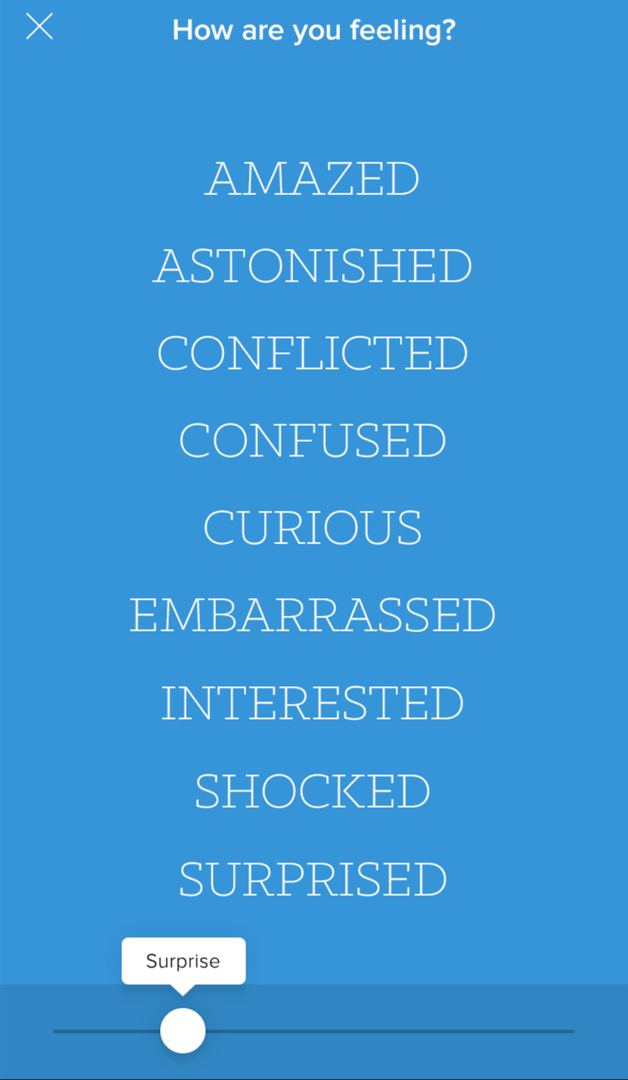}
    \caption{Emotion Picker}
    \label{fig:VentEmoPicker}
  \end{subfigure}
  \caption{Screenshots of the interface of Vent app, taken from~\cite{lykousas2019}.}
  \label{fig:VentScreenshot}
\end{figure}

\begin{figure}[!ht]
    \centering
    \includegraphics[width=1.0\linewidth]{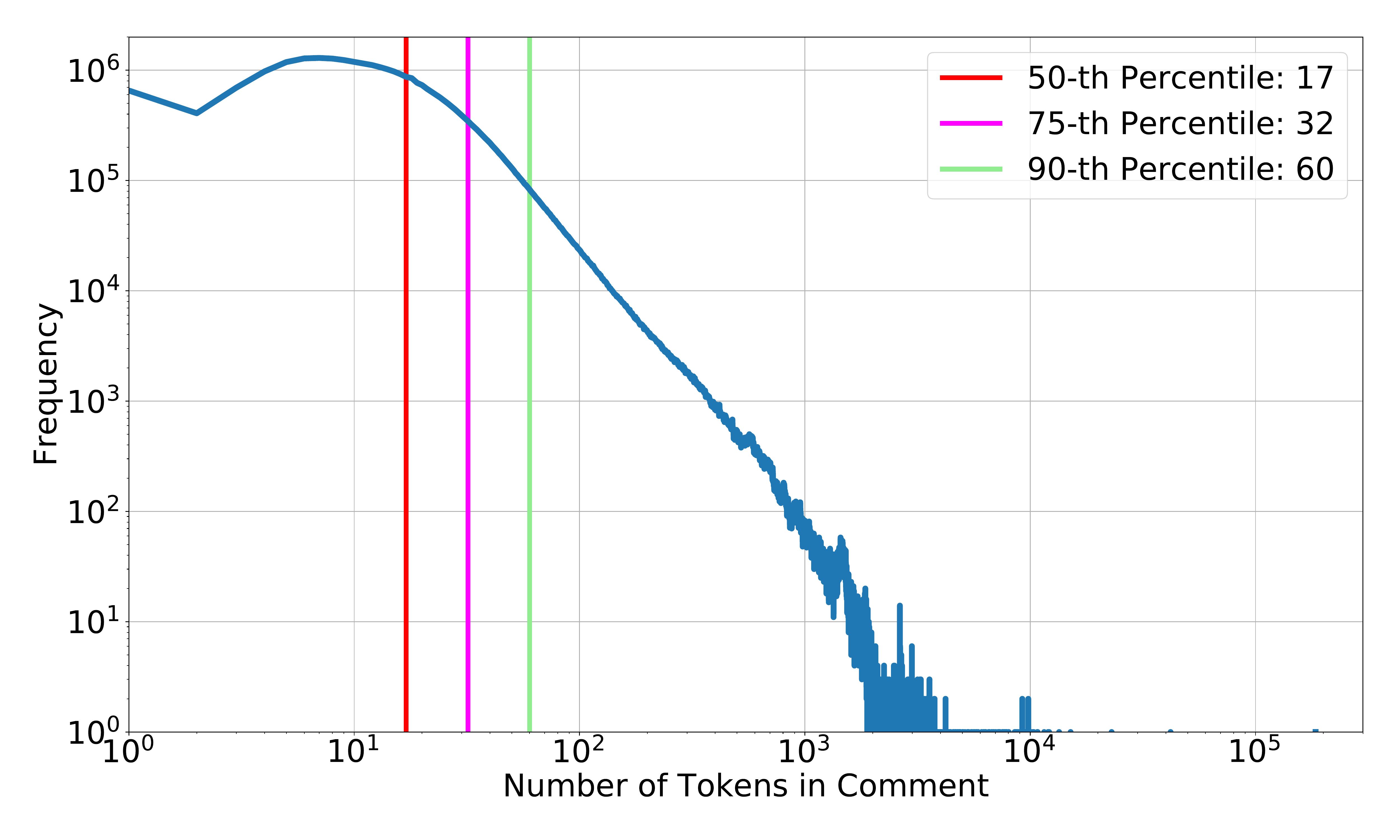}
    \caption{Vent comment length distribution in number of tokens. }
    \label{fig:VentTokenLengthDistribution}
\end{figure}

The Vent dataset is self-annotated by Vent users (writers) using the interfaces shown in~\autoref{fig:VentScreenshot}. The distribution of comment lengths in number of tokens is shown in~\autoref{fig:VentTokenLengthDistribution}. We used 32 as the maximum number of tokens per Vent in alignment with the $75$\textsuperscript{th} percentile in the length distribution. 

\subsubsection{Filtering procedure}

We filter the dataset to (a) contain months with sufficient volume and (b) only contain stable emotions that appear every month onwards from the chosen point. 

\begin{figure*}[!ht]
\centering
  \centering
  \includegraphics[width=1.0\linewidth]{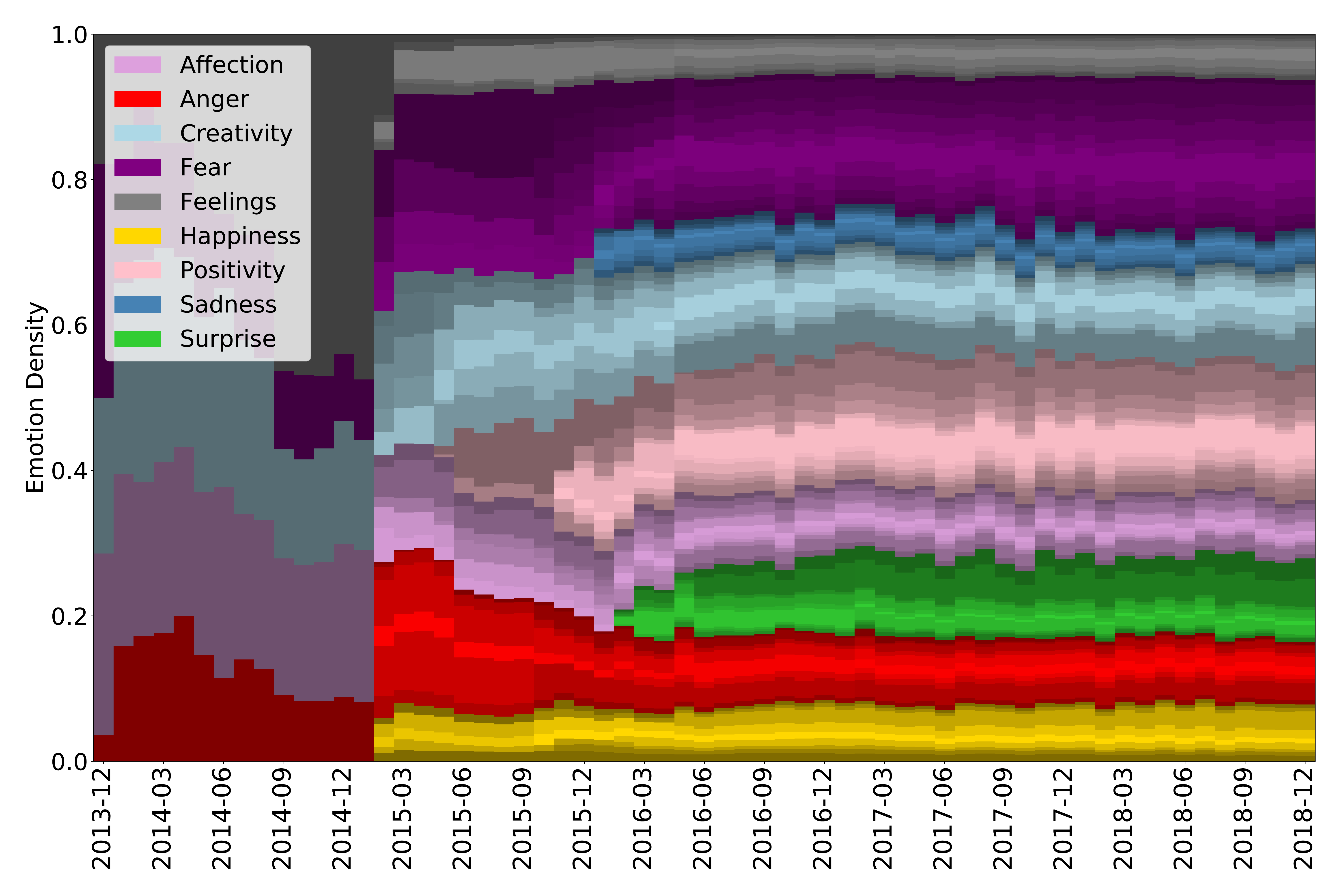}
  \caption{Category-Level densities on a month-by-month basis. Each color represents an individual emotion, with its total size being the percentage of the emotion across all messages in that month. Densities are irregular initially, until the number of emotions in the app plateaus and all categories become defined.}
  \label{fig:VentEmotionLandscape}
\end{figure*}

In~\autoref{fig:VentEmotionLandscape}, we show the distribution of emotions over time in terms of the overall frequency of emotions from each of the 9 emotion categories. 
In~\cite{lykousas2019}, the authors report user activity peaks in April 2015, and slowly decreases afterwards. 

\begin{figure*}[!ht]
\centering
  \centering
  \includegraphics[width=0.96\linewidth]{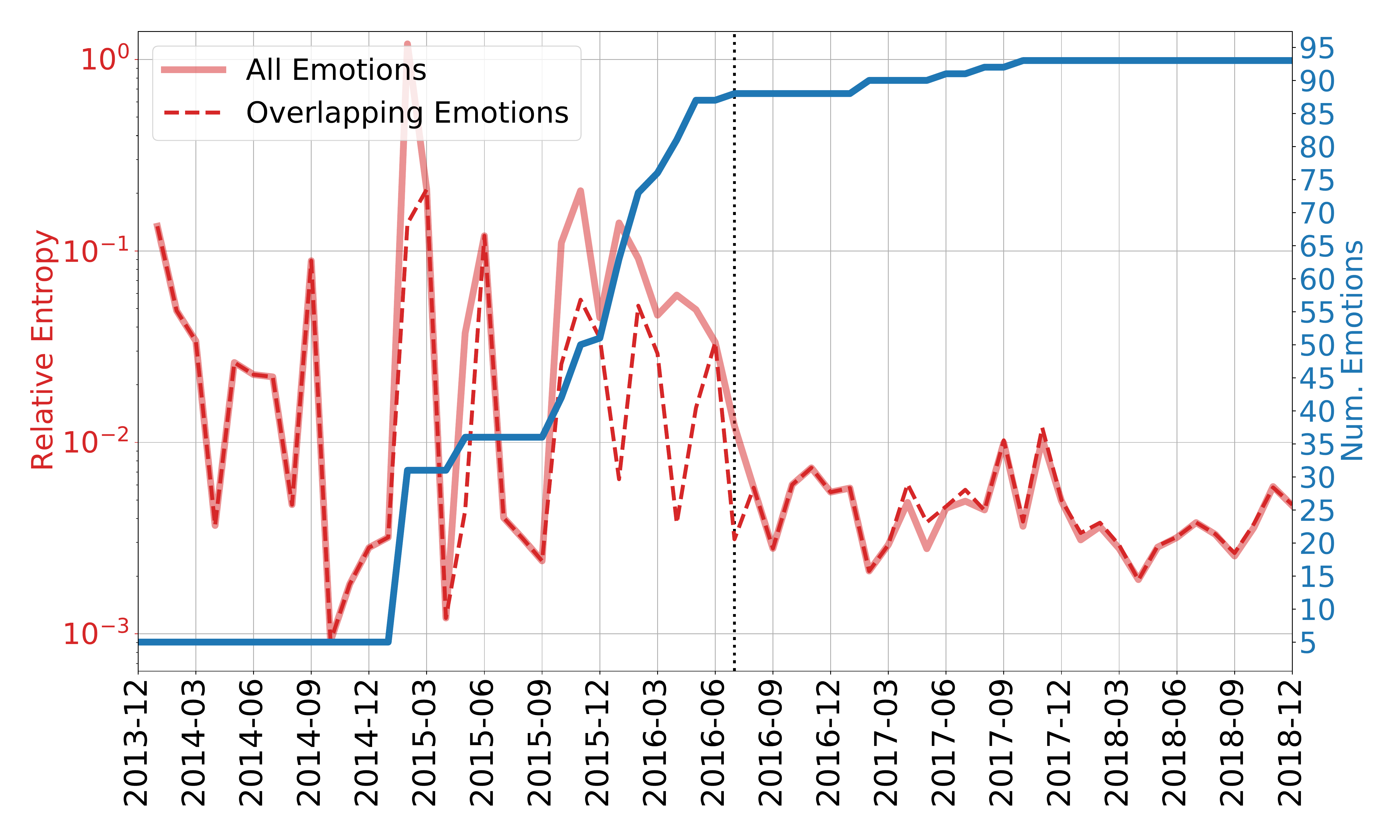}
  \caption{Relative Entropies and Distinct Emotion Counts month over month. The relative entropy of \emph{all} emotions includes peaks when new emotions are released, while overlapping emotions capture the stability of emotions that persist between two months. The number of emotions grows over time as new emotions are released and replaced. We find that the number of emotions and the relative entropies stabilize after 2016-07, represented by the dashed mark.}
  \label{fig:StableRelativeEntropies}
\end{figure*}

We select the filter in July 2016 according to two criteria: first, we compute the number of valid distinct emotions over time (see main text for definition of a valid emotion). This is shown in~\autoref{fig:StableRelativeEntropies} (blue line). We observe that after July 2016, the large majority of distinct emotions have been created. Second, to characterize their stability, we compute the relative entropy between the emotion distributions of two consecutive months. This is shown in \autoref{fig:StableRelativeEntropies} (red lines). 
The emotion distributions consider the intersection (dashed red) or union (solid red) of emotions in each pair of consecutive months. We observe that after our cut-off in July 2016 both curves show that relative entropy takes regular values below $0.02$, indicating that the use of emotions is stable after that month.

\subsubsection{Obscene Word Analysis}

Unlike other works like GoEmotions~\cite{demszky2020}, we do not exclude content with foul language from the training process. Our rationale is that Vent is a dataset collected from a moderated social network. In this sense, we assume moderators remove toxic or unacceptable behaviour, while hypothesizing that foul language is a strong emotional signal that is necessary to capture emotional messages in certain categories. We note, however, that we do filter \texttt{nsfw} (not safe for work) content during the human annotation task to avoid exposing readers to inappropriate content.

In order to justify the decision to keep obscene terms, we study whether there are significant differences in the degree of obscenity for each emotion and category. We flag whether or not a comment is obscene by detecting 763 words from two publicly available obscene~{\footnote{\url{https://github.com/RobertJGabriel/Google-profanity-words/blob/master/list.txt}}} and bad word~{\footnote{\url{https://code.google.com/archive/p/badwordslist/downloads}}} lists. We exclude frequency analysis from our study: rather than accounting for obscene word frequencies, we use a binary indicator that signals whether an obscene word from either list is observed in a Vent comment.

\begin{figure*}[!ht]
    \centering
    \includegraphics[width=0.9\linewidth]{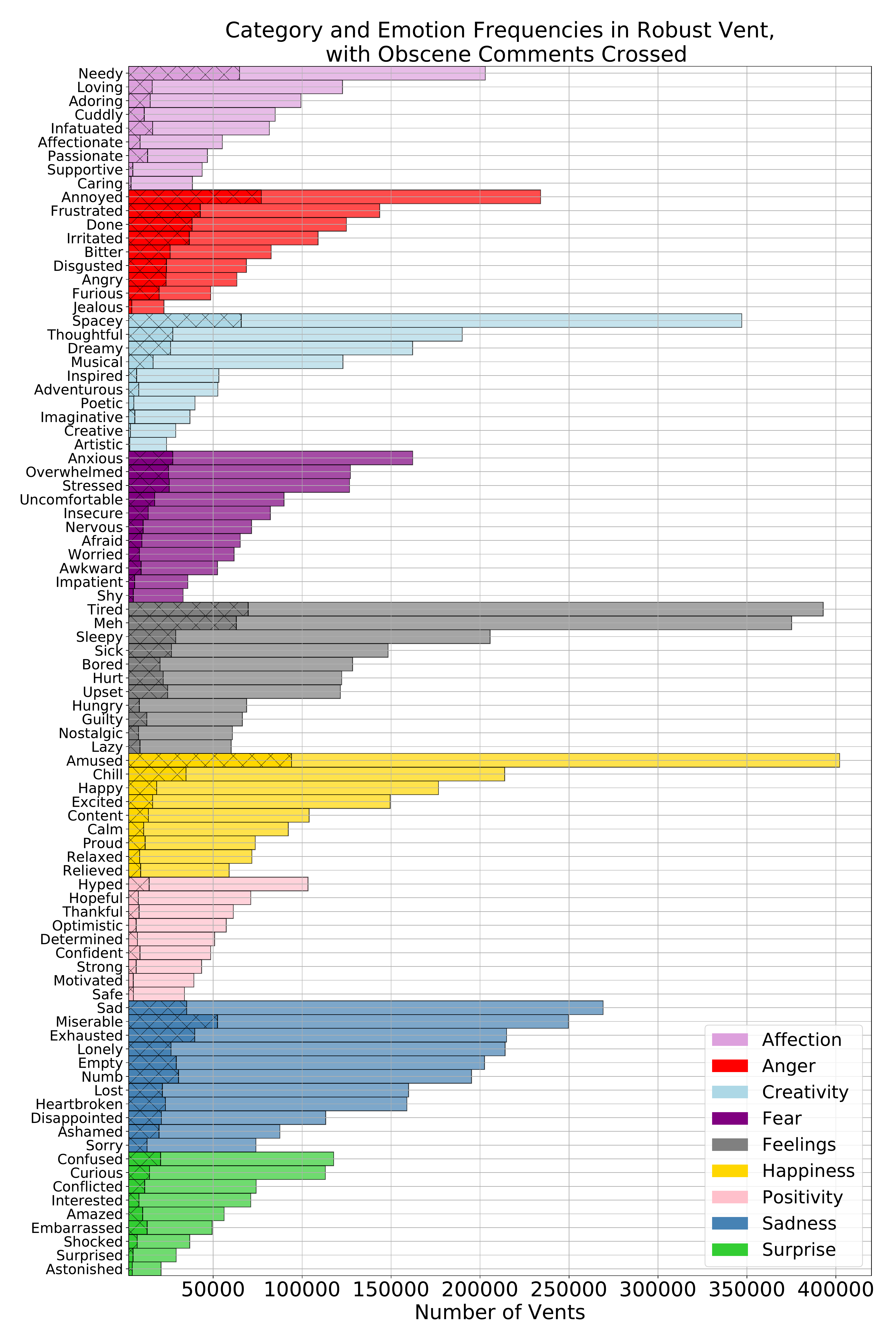}
    \caption{Vent emotion labels and their frequencies after filtering for relevant emotions. Crossed darker regions represent the proportion of a given category that contains at least one obscene term. The colors showcase the emotion categories, which approximately align with Ekman's: \emph{happiness, fear, sadness, surprise} and \emph{anger} are present, while \emph{disgust} is treated as an emotion under \emph{anger}. There are four additional categories that match higher-level subjective states: \emph{creativity, positivity, feelings} and \emph{affection}.}
    \label{fig:VentDataCharacteristics}
\end{figure*}

We observe significant differences ($p < 10^{-20}$, using bootstrapped z-tests) in the percentage of vents containing obscene words between categories, estimated with bootstrapping on 100 independent runs with 10\% of the dataset sampled with replacement. We find that 10\% comments in the Positive category contain obscene words (lowest) while 32\% of Anger vents use obscene language (highest). \autoref{fig:VentDataCharacteristics} provides an overview of the number of snippets per emotion and category, and the proportion of obscene comments for each emotion. 

\section{Benchmark Design}

\subsection{Representation Approaches}

We implement statistical and embedded representations. In particular, we employ 4 different algorithms:

\begin{asparaenum}
    \item \textbf{Bag-of-Words.} Each document is represented as a sparse vector containing the counts of every word. We limit the vocabulary size after filtering English stop words using Scikit-learn's list~\cite{scikit-learn} but without applying any transformations like stemming. Our vocabulary contains only the most frequent words, which we tune as a hyper-parameter shown in~\autoref{tab:HyperParameters}.
    \item \textbf{TF-IDF.} Each document is represented as a sparse vector containing the normalized term-frequency divided by inverse document frequency. We apply the same vocabulary-building policy as on Bag-of-Words, including stop-word filtering and using the top most frequent words tuned as a hyper-parameter shown in~\autoref{tab:HyperParameters}.
    \item \textbf{FastText.} We use a pre-trained unsupervised English FastText~\cite{bojanowski2017} model to embed the sequence of tokens in a sentence. We limit the length of the sentence as a hyper-parameter.
    \item \textbf{BERT.} We use a pre-trained BERT~\cite{devlin2019} English model to tokenise and embed the sequence of tokens in a sentence. Our design permits gradients to propagate when BERT is used as an input to a downstream neural model. We limit the length in total tokens and whether or not to propagate gradients (e.g. freeze BERT) as hyper-parameters shown in~\autoref{tab:HyperParameters}.
\end{asparaenum}

\subsection{Modeling Methods}

We employ five different learners:
\begin{asparaenum}
    \item \textbf{Naive Bayes.} We employ Scikit-learn~\cite{scikit-learn}'s implementation of Multinomial Naive Bayes in a one-vs-rest setting for multi-label classification.
    \item \textbf{Logistic Regression.} Likewise, we use Scikit-learn's SGD-based Logistic Regression in a one-vs-rest setting for multi-label classification.
    \item \textbf{Incremental Random Forests.} We use an extension to Scikit-learn's Random Forest classifier to incrementally build the forest in a mini-batch fashion, using the \texttt{incremental-trees} Python library~\cite{incremental-trees}. Each mini-batch grows a fixed number of trees, up to a maximum number, for all batches.
    \item \textbf{(Pooled) Neural Networks.} We implement a simple DNN architecture, applied to every embedded token in parallel with $N$ output neurons, as many as there are classes. The final result is computed by applying a pooling function over the states, as shown in \autoref{fig:DNNPoolArchitecture}.
    \item \textbf{(Bi)-LSTM.} We implement a stacked (Bi)-LSTM architecture that consumes an embedded sequence in a seq2seq manner, optionally processing in a single direction. The recurrent network captures inter-token relationships and produces $N$ output neurons per token. The final result is computed by applying a pooling function over each token, as shown in \autoref{fig:BiLSTMPoolArchitecture}.
\end{asparaenum}

\section{Hyper-parameter Configuration}

All the hyper-parameters used when tuning the algorithms used in our experiments are shown in~\autoref{tab:HyperParameters}. Each row shows a hyper-parameter affecting either the whole training process, the representation method, or the learning algorithm. Values in \textbf{bold} were chosen through Grid Search as the best performing values for that particular component and data set. Our grid search optimizes micro-F1 for each method, running on a Slurm cluster with 8GB Nvidia GPUs. Each configuration was executed once to identify candidates, with the final experiments executing the best performing configuration on 5 independent runs using different seeds. In the code samples, we provide the original \texttt{.json} files containing the exact run configurations for reproducibility, including execution files that contain the per-run evaluation metrics to trace the notebooks used to compute the values reported in the paper.

\begin{table*}[!ht]
\centering
\tiny
\begin{tabular}{@{}llll@{}}
\toprule
\multicolumn{1}{c}{\normalsize\textbf{Dataset}} & \multicolumn{1}{c}{\normalsize\textbf{Component}} & \multicolumn{1}{c}{\normalsize\textbf{Hyper-parameter}} & \multicolumn{1}{c}{\normalsize\textbf{Values}}                                                                \\ \midrule
\multirow{32}{*}{\normalsize{GoEmotions}}          & \normalsize{Global}                                 & Batch Size                                   & \begin{tabular}[c]{@{}l@{}}100000 (statistical methods),\\ 1024 (FastText), 64 (BERT)\end{tabular} \\ \cmidrule(l){2-4} 
                                      & \normalsize{Bag-of-Words}                           & Vocabulary Size                              & \textbf{5000}, 10000, 20000, 40000                                                               \\ \cmidrule(l){2-4} 
                                      & \normalsize{TF-IDF}                                 & Vocabulary Size                              & \textbf{5000}, 10000, 20000, 40000                                                           \\ \cmidrule(l){2-4} 
                                      & \multirow{3}{*}{\normalsize{BERT}}                  & Freeze                                       & True, \textbf{False}                                                                                    \\
                                      &                                        & Model                                        & bert-base-cased, \textbf{bert-base-uncased}                                                             \\
                                      &                                        & Max Length                                   & 25                                                                                                 \\ \cmidrule(l){2-4} 
                                      & \multirow{2}{*}{\normalsize{FastText}}              & Model                                        & Common Crawl English Model                                                                         \\
                                      &                                        & Max Length                                   & 25                                                                                                 \\ \cmidrule(l){2-4} 
                                      & \multirow{3}{*}{\normalsize{Log. Reg.}}             & Epochs                                       & 1, 10, 50, \textbf{100}                                                                                 \\
                                      &                                        & $\alpha$                        & 0.00001, \textbf{0.0001}, 0.001, 0.01, 0.1, 1.0                                                         \\
                                      &                                        & Tolerance                                    & 0.001                                                                                              \\ \cmidrule(l){2-4} 
                                      & \normalsize{Naive Bayes}                            & Smoothing Factor                             & 1, \textbf{0.1}, 0.01, 0.001, ..., 1e-10                                                                \\ \cmidrule(l){2-4} 
                                      & \multirow{4}{*}{\normalsize{Random Forest}}         & Trees per Batch                              & \textbf{1000}, 2000, 3000                                                                               \\
                                      &                                        & Max. Depth                                   & 3, 4, 5, 6, \textbf{7}                                                                                  \\
                                      &                                        & Max. Features Fraction                       & \textbf{0.05}, 0.1, 0.2, 0.4                                                                            \\
                                      &                                        & Split Criterion                              & Entropy                                                                                            \\ \cmidrule(l){2-4} 
                                      & \multirow{8}{*}{\normalsize{DNN Pool}}              & Hidden Size                                  & 100                                                                                                \\
                                      &                                        & Num. Layers                                  & 1, \textbf{2}, 3                                                                                        \\
                                      &                                        & Num. Epochs                                  & \textbf{30}, 40, 50, 60                                                                                 \\
                                      &                                        & Learning Rate                                & 0.01, 0.001, \textbf{0.0001}                                                                            \\
                                      &                                        & Epsilon                                      & 1e-5, \textbf{1e-6}, 1e-7                                                                               \\
                                      &                                        & Activation                                   & ELU, \textbf{Tanh}                                                                                      \\
                                      &                                        & Pooling Function                             & Attention, Mean, \textbf{Max}                                                                           \\
                                      &                                        & Optimizer                                    & AdamW                                                                                              \\ \cmidrule(l){2-4} 
                                      & \multirow{8}{*}{\normalsize{Bi-LSTM}}               & Hidden Size                                  & 100                                                                                                \\
                                      &                                        & Num. Layers                                  & 1, \textbf{2}                                                                                           \\
                                      &                                        & Num. Epochs                                  & \textbf{30}, 40, 50, 60                                                                                 \\
                                      &                                        & Learning Rate                                & \textbf{0.01}, 0.001, 0.0001                                                                            \\
                                      &                                        & Epsilon                                      & \textbf{1e-5}, 1e-6, 1e-7                                                                               \\
                                      &                                        & Bidirectional                                & \textbf{True}, False                                                                                    \\
                                      &                                        & Pooling Function                             & Attention, \textbf{Mean}, Max                                                                           \\
                                      &                                        & Optimizer                                    & AdamW                                                                                              \\ \midrule
\multirow{32}{*}{\normalsize{Vent}}                & \normalsize{Global}                                 & Batch Size                                   & \begin{tabular}[c]{@{}l@{}}100000 (statistical methods),\\ 2048 (FastText), 64 (BERT)\end{tabular} \\ \cmidrule(l){2-4} 
                                      & \normalsize{Bag-of-Words}                           & Vocabulary Size                              & \textbf{5000}, 10000, 20000, 40000, 80000                                                         \\ \cmidrule(l){2-4} 
                                      & \normalsize{TF-IDF}                                 & Vocabulary Size                              & \textbf{5000}, 10000, 20000, 40000, 80000                                                         \\ \cmidrule(l){2-4} 
                                      & \multirow{3}{*}{\normalsize{BERT}}                  & Freeze                                       & True, \textbf{False}                                                                                    \\
                                      &                                        & Model                                        & bert-base-cased, \textbf{bert-base-uncased}                                                             \\
                                      &                                        & Max Length                                   & 40                                                                                                 \\ \cmidrule(l){2-4} 
                                      & \multirow{2}{*}{\normalsize{FastText}}              & Model                                        & Common Crawl English Model                                                                         \\
                                      &                                        & Max Length                                   & 40                                                                                                 \\ \cmidrule(l){2-4} 
                                      & \multirow{3}{*}{\normalsize{Log. Reg.}}             & Epochs                                       & 1, 2, 10, \textbf{50}, 100                                                                              \\
                                      &                                        & $\alpha$                        & \textbf{0.00001}, 0.0001, 0.001, 0.01, 0.1, 1.0                                                         \\
                                      &                                        & Tolerance                                    & 0.001                                                                                              \\ \cmidrule(l){2-4} 
                                      & \normalsize{Naive Bayes}                            & Smoothing Factor                             & 1, 0.1, \textbf{0.01}, 0.001, ..., 1e-10                                                                \\ \cmidrule(l){2-4} 
                                      & \multirow{4}{*}{\normalsize{Random Forest}}         & Trees per Batch                              & \textbf{66}, 125, 250, 500                                                                              \\
                                      &                                        & Max. Depth                                   & 4, \textbf{5}                                                                                           \\
                                      &                                        & Max. Features Fraction                       & \textbf{0.05}, 0.1, 0.2, 0.4                                                                            \\
                                      &                                        & Split Criterion                              & Entropy                                                                                            \\ \cmidrule(l){2-4} 
                                      & \multirow{8}{*}{\normalsize{DNN Pool}}              & Hidden Size                                  & \textbf{100}, 200                                                                                       \\
                                      &                                        & Num. Layers                                  & 1, \textbf{2}, 3, 4, 5                                                                                  \\
                                      &                                        & Num. Epochs                                  & 1, 2, \textbf{3}                                                                                        \\
                                      &                                        & Learning Rate                                & \textbf{0.01}, 0.001, 0.0001                                                                            \\
                                      &                                        & Epsilon                                      & \textbf{1e-5}, 1e-6, 1e-7                                                                               \\
                                      &                                        & Activation                                   & \textbf{ELU}, Tanh                                                                                      \\
                                      &                                        & Pooling Function                             & \textbf{Attention}, Mean, Max                                                                           \\
                                      &                                        & Optimizer                                    & AdamW                                                                                              \\ \cmidrule(l){2-4} 
                                      & \multirow{8}{*}{\normalsize{Bi-LSTM}}               & Hidden Size                                  & 100, \textbf{200}, 400                                                                                  \\
                                      &                                        & Num. Layers                                  & 1, \textbf{2}, 3                                                                                        \\
                                      &                                        & Num. Epochs                                  & 1, \textbf{2}                                                                                           \\
                                      &                                        & Learning Rate                                & 0.01, \textbf{0.001}, 0.0001                                                                            \\
                                      &                                        & Epsilon                                      & \textbf{1e-5}, 1e-6, 1e-7                                                                               \\
                                      &                                        & Pooling Function                             & \textbf{Attention}, Mean, Max                                                                           \\
                                      &                                        & Bidirectional                                & \textbf{True}, False                                                                                    \\
                                      &                                        & Optimizer                                    & AdamW                                                                                              \\ \bottomrule
\end{tabular}
\caption{Hyper-parameters used by the different representation and modelling algorithms tested in our benchmark. Each hyper-parameter configuration is tested once to find candidate configurations with micro-F1 on the validation set as our objective. In \textbf{bold}, the best performing hyper-parameter value out of all configurations using a specific method.}
\label{tab:HyperParameters}
\vspace{1cm}
\end{table*}

\section{Experimental Setting}

We perform 5 independent experiment runs using different seeds for all underlying libraries used in our implementation. For every experiment configuration, we report the average performance across the 5 runs computed for each metric.

\section{Experiment Results}

\begin{figure*}[!ht]
    \centering
    \includegraphics[width=1.0\linewidth]{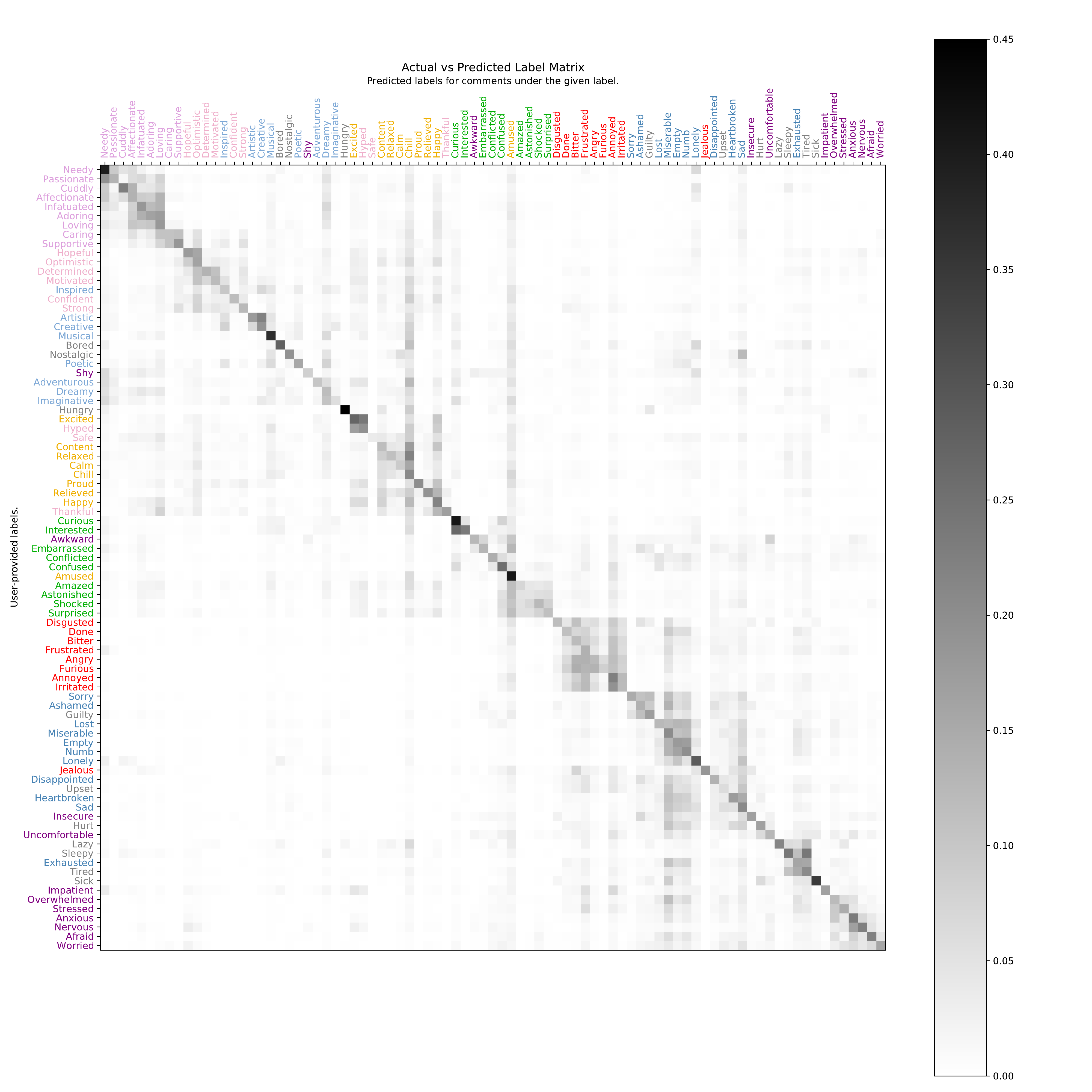}
    \caption{Vent emotion confusion matrix for the best performing model (BERT + Bi-LSTM).}
    \label{fig:VentEmotionConfusionMatrix}
\end{figure*}

We report per-category performances beyond the results reported in the main text in~\autoref{tab:VentCategoryF1}, and show the confusion matrix for all emotions in~\autoref{fig:VentEmotionConfusionMatrix}.

\begin{table}[ht]
\centering
\begin{tabular}{@{}l@{\hskip 1pt}rrr@{}}
\toprule
\textbf{Category}   & \textbf{Precision} & \textbf{Recall} & \textbf{F1-Score} \\ \midrule
\textbf{Affection}  & \textbf{0.51}     & 0.48           & 0.49             \\
\textbf{Anger}      & 0.42              & 0.37           & 0.40             \\
\textbf{Creativity} & \textit{0.25}     & 0.41           & 0.31             \\
\textbf{Fear}       & 0.39              & 0.36           & 0.35             \\
\textbf{Feelings}   & 0.33              & 0.45           & 0.38             \\
\textbf{Happiness}  & 0.41              & 0.45           & 0.43             \\
\textbf{Positivity} & 0.29              & \textit{0.26}  & \textit{0.27}             \\
\textbf{Sadness}    & \textbf{0.51}     & \textbf{0.52}  & \textbf{0.51}             \\
\textbf{Surprise}   & 0.30              & 0.35           & 0.33             \\ \bottomrule
\end{tabular}
\caption{Per-category performance of the best performing model (BERT / Bi-LSTM) on Vent. In \textbf{bold}, the best performing category for each metric; in \emph{cursive}, the worst performing category.}
\label{tab:VentCategoryF1}
\end{table}

\subsection{Model Performance vs Dataset Size}
\begin{figure*}[!ht]
    \centering
    \includegraphics[width=1.05\linewidth]{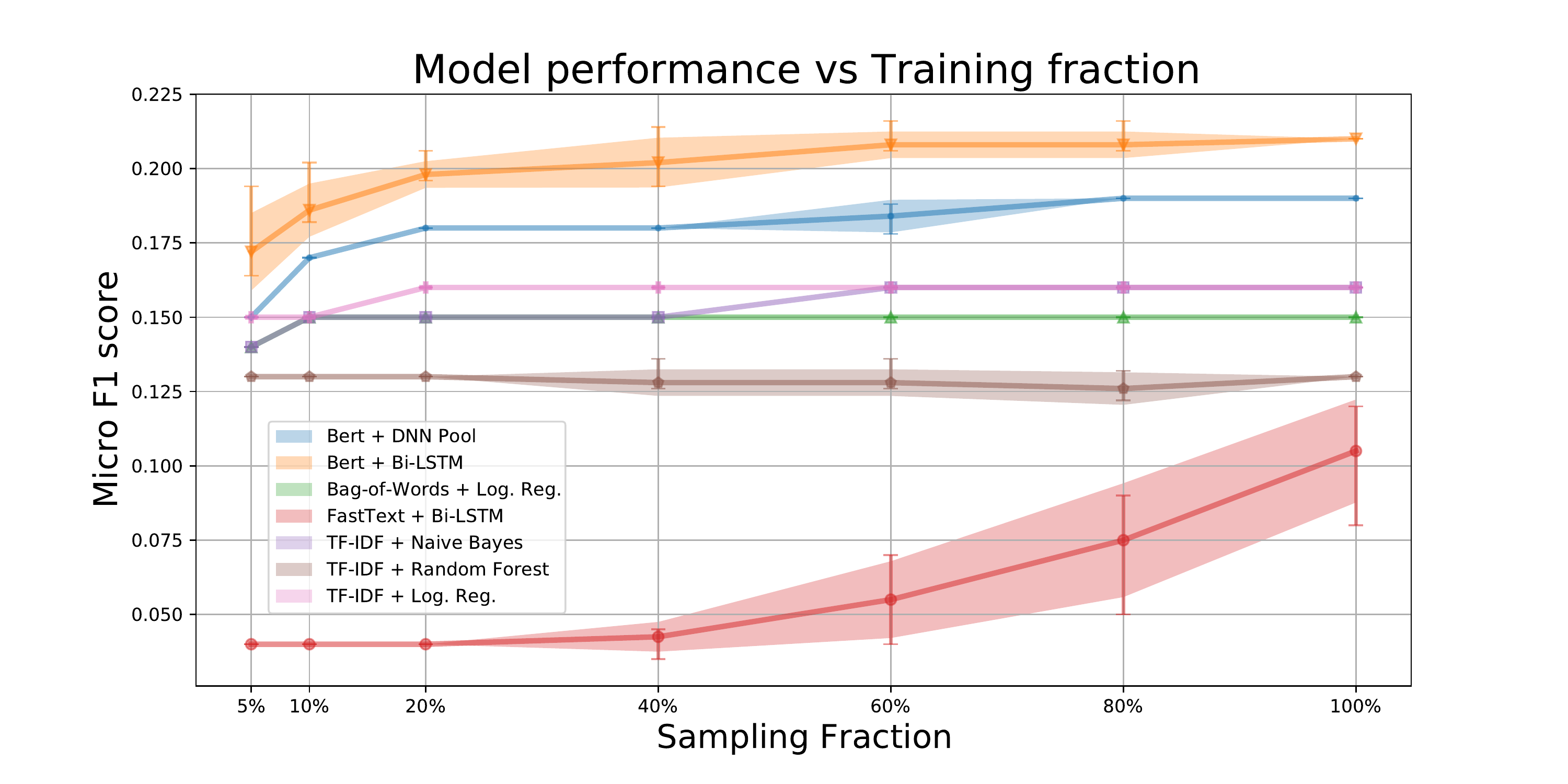}
    \caption{Model performance (micro-F1 score) against the whole test set given a percentage of training data.}
    \label{fig:VentModelPerfTrainingFraction}
\end{figure*}

To check the performance that our models as a function of the data size, we analyze how models perform under different fractions of the Vent dataset. 
%We study how models perform under different fractions of the Vent dataset. This serves as a sanity check to ensure that our models are reasonable learners capable of better capturing the hypothesis when provided with more data. It also provides some insight on the performance curves of the different algorithms in the benchmark. 
\autoref{fig:VentModelPerfTrainingFraction} shows the micro-F1 score curves for given percentage of data available.

We find that statistical models show stable performance, which we explain by the fact that the vocabulary sizes are chosen by searching on a limited set of values. We also observe that FastText slowly improves but requires a much higher data intensity than other methods, which we believe is caused by the inability of re-training the FastText model through backpropagation due to limitations in the implementation.

\subsection{Temporal Impact Emotional Data}

We compare whether training with Vent using uniformly random splits to study whether the data is non-stationary. We take the Robust subset of Vent and divide randomly it in 3 splits whose sizes match the previous 80 / 10 / 10 setting. Our findings show consistently worse results, as observed in \autoref{tab:VentRandomBenchmark}. For instance, the performance of the BERT / Bi-LSTM model in terms of micro-F1 is 0.19, against the 0.21 observed for the model trained with temporal splits which we discuss in the main body of the paper. We believe this behavior might be caused by the homophilic nature of social networks: as the network grows, new members join existing communities and over time their vocabularies homogenize, becoming more predictable over time. This would be consistent with studies on the socio-linguistic evolution of communities~\cite{danescu2013}, which predict the duration of the life span of a user in a community given their posting behaviour---with users that do not align with the linguistic expectations of the community deciding to leave, and those that fall in line reinforcing their cultural norms. 

\begin{table}[!ht]
\centering
\begin{tabular}{@{}l@{\hskip 3pt}l@{}r@{\hskip 3pt}r@{\hskip 3pt}r@{\hskip 3pt}r@{}}
\toprule
\multicolumn{1}{@{}c}{\textbf{Repr.}} & \multicolumn{1}{c}{\textbf{Model}} & \multicolumn{1}{@{}c@{}}{\textbf{M-F1}} & \multicolumn{1}{c@{}}{\textbf{m-F1}} & \multicolumn{1}{c}{\textbf{Pre}} & \multicolumn{1}{c@{}}{\textbf{Rec}} \\ \midrule
--- & Random & 0.02 & 0.04 & 0.02 & 0.60 \\\midrule
\multirow{3}{*}{BoW} & N. Bayes & 0.13 & 0.15 & 0.12 & 0.22 \\
                     & Log. Reg. & 0.13 & 0.15 & 0.13 & 0.20 \\
                     & R. Forest & 0.11 & 0.13 & 0.12 & 0.17 \\\midrule
\multirow{3}{*}{TF-IDF} & N. Bayes & 0.14 & 0.16 & 0.13 & 0.20 \\
                        & Log. Reg. & 0.14 & 0.16 & 0.14 & 0.20 \\
                        & R. Forest & 0.11 & 0.13 & 0.12 & 0.17 \\\midrule
\multirow{2}{*}{FT} & DNN Pool & 0.10 & 0.13 & 0.10 & 0.21 \\
                    & Bi-LSTM & 0.04 & 0.05 & 0.03 & \textbf{0.32} \\\midrule
\multirow{2}{*}{BERT} & DNN Pool & 0.16 & 0.18 & 0.16 & 0.23 \\
                      & Bi-LSTM & \textbf{0.18} & \textbf{0.19} & \textbf{0.17} & 0.24 \\\bottomrule
\end{tabular}
\caption{Averaged results on the Vent dataset with Random Splits. Best performing models are in \textbf{bold}. The metrics are \textbf{M}acro F1, \textbf{m}icro F1, and micro-averaged \textbf{Pre}cision and \textbf{Rec}all. All standard deviations $\leq$ 0.01.}
\label{tab:VentRandomBenchmark}
\end{table}

\subsection{Emotional Transfer Learning}

Vent is a large dataset which we expect to help specialise BERT for emotion recognition tasks. We expect that the resulting GoEmotions model will perform better (as measured by micro F1-Score) than the model trained directly on GoEmotions, as the last layers of BERT will have been fine-tuned with the emotional content of Vent. %\AKn{Add maybe a figure to show this idea and the corresponding results in table} \DNn{Results in the table correspond to a previous experiment, we do not have significant numbers for transfer learning.}

To measure the amount of emotional information in Vent, we implement a transfer learning task with GoEmotions. We train model with the best performing BERT / Bi-LSTM configuration on different Vent subsets. For every trained model, we take the finetuned BERT embedder and use it as the seed embedding model to repeat our experiments GoEmotions changing no other hyper-parameter. No improvements are found on the transfer learning task (results not shown), so we believe that the signal is encoded in the task-specific Bi-LSTM model rather than within the BERT layers. 

\subsection{Human Reader Evaluation}

We provide additional details on the design of the HIT, experimental results, and our compensation structure to ensure that workers on our HITs receive a fair compensation.

\subsubsection{Annotation Procedure}

\begin{figure*}[!ht]
    \centering
    \includegraphics[width=0.95\linewidth]{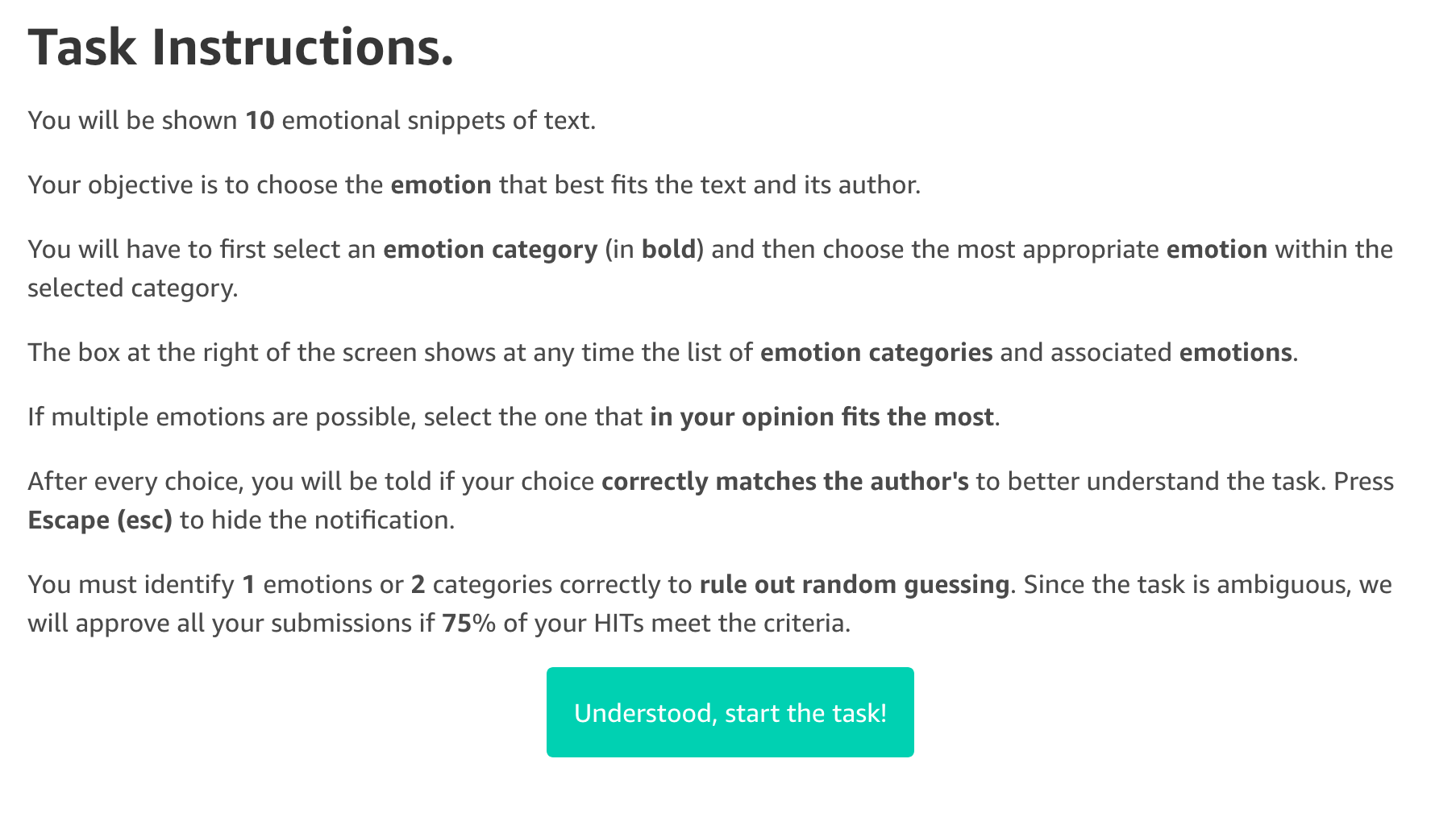}
    \caption{Instructions shown to the MTurk workers.}
    \label{fig:MTurkInstructions}
\end{figure*}

\begin{figure*}[!ht]
  \centering
  \includegraphics[width=1.0\linewidth]{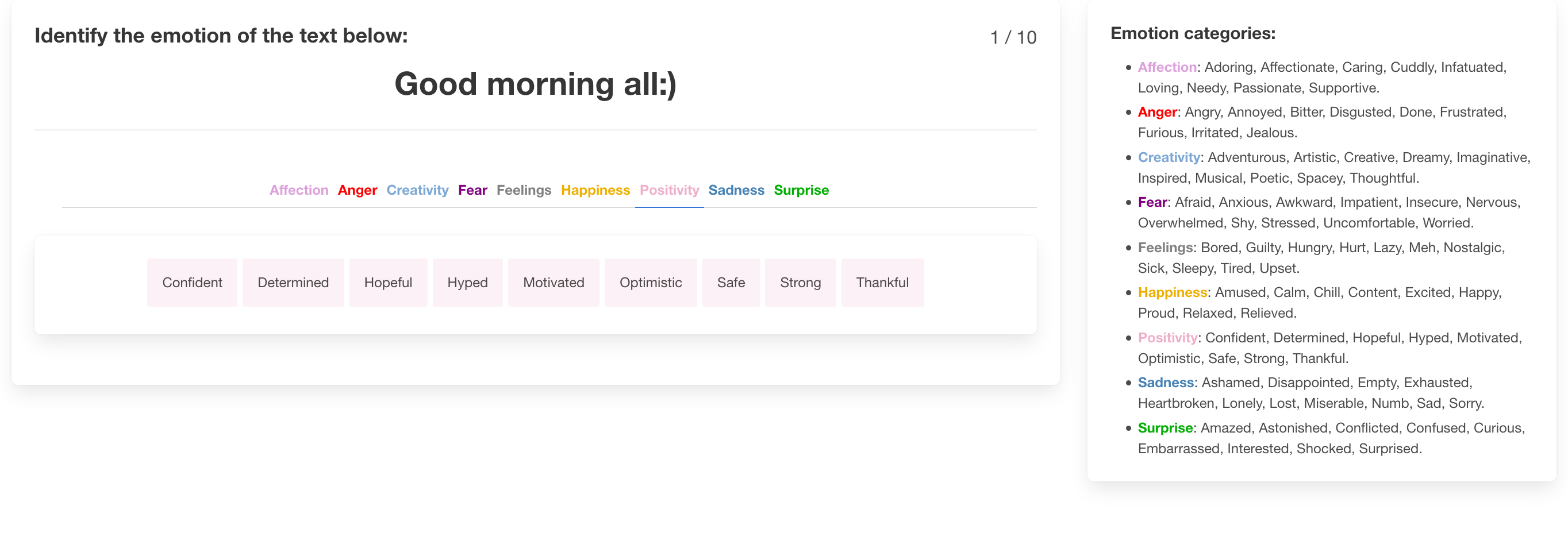}
  \caption{Screenshot of the MTurk Annotation Tool. Workers must first select an emotion category, then choose the emotion within that category they find more appropriate.}
  \label{fig:MTurkAnnotationScreenshot}
\end{figure*}

Workers are tasked to annotate the emotions for 10 comments, whose order is shuffled in every distinct session to avoid position biases. The annotation workflow starts with a set of instructions,  shown in \autoref{fig:MTurkInstructions}, emphasizing they have to match the author's (writer) emotion. Then the task starts showing texts, as illustrated in~\autoref{fig:MTurkAnnotationScreenshot}. Workers first select an emotion category and then choose the emotion according to the previous instructions. Upon selecting an emotion, a message prompts the worker on whether the prediction was correct or not at the emotion and category level, given the original Vent ground truth. %We introduce this feedback mechanism to help workers understand the task in terms of \textit{writer} emotion detection, that is, predicting what the writer could\AKn{This is important, we should say that in the main article} have felt.

To ensure that workers submit quality work, and that the assessment is equal for all workers, we define approval rules based on their predictive performance. For any submission, we expect that it is distinguishable from random choices, which means that it must contain correct predictions for at least 1 out of 88 emotions or 2 out of 9 emotion categories. We implement both constraints as an automated check of annotator quality, accepting or rejecting the provided annotations for every new submission.

\begin{figure}[!ht]
    \centering
    \includegraphics[width=1.0\linewidth]{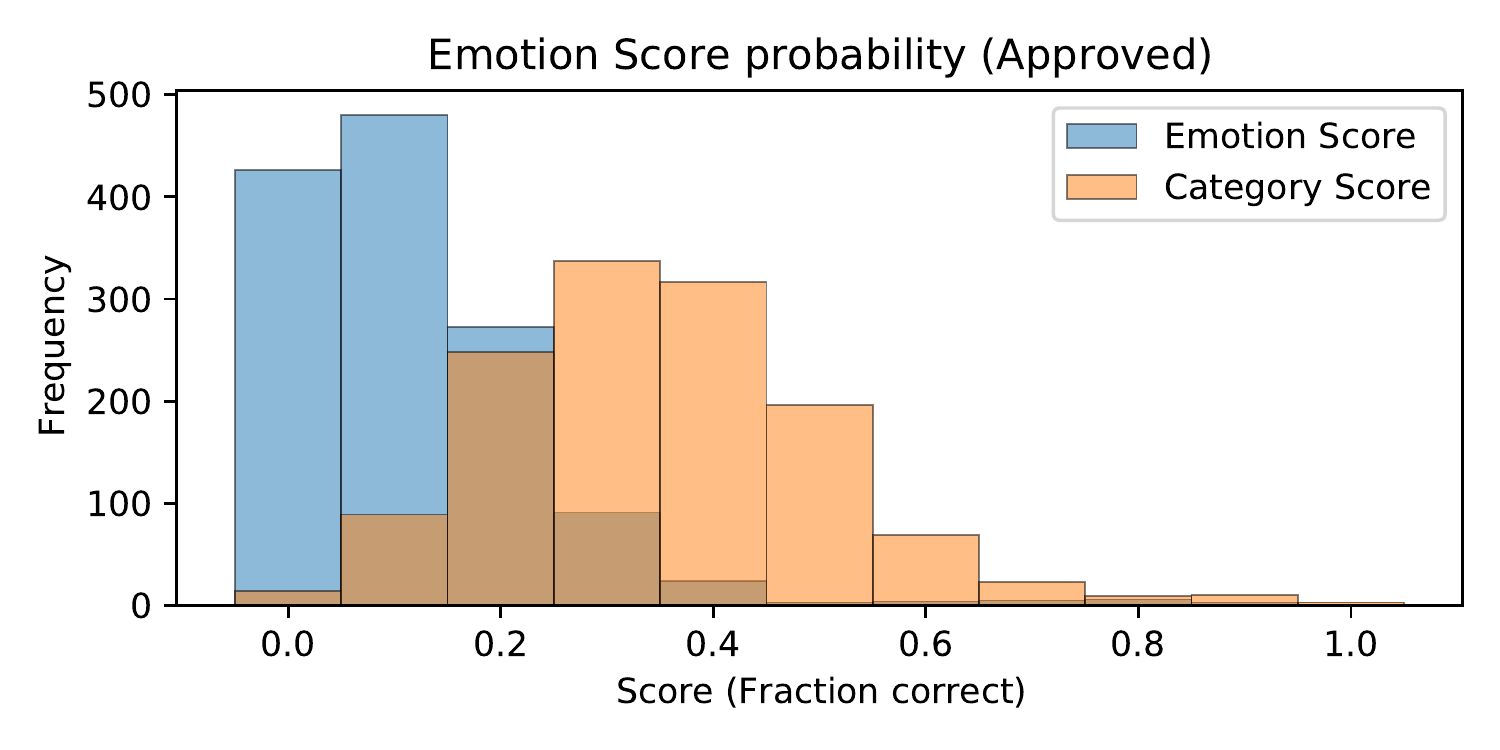}
    \caption{Emotion and category accuracy scores for the readers, computed as the accuracy of the reader against the author-provided label on every task.}
    \label{fig:MTurkEmotionScores}
\end{figure}

In order to account for the ambiguity of the task and avoid penalising quality workers\footnote{During our trial run (whose data we do not use in our final study), workers contacted us about the ambiguity and meaninglessness of the emotions provided by Vent users and the impact of rejections in their future earnings on the platform.}, we approve all tasks submitted by workers that meet the quality criteria in 75\% of their work. By the end of our 264 HIT batch, we approved 97.33\% of the submissions sent by 84 different readers with an average emotion accuracy of 11.43\% and an average category accuracy of 34.26\%. The distribution of accuracy scores is shown in \autoref{fig:MTurkEmotionScores}.

%\say{
%You will be shown 10 emotional snippets of text.
%Your objective is to choose the emotion that best fits the text and its author.
%You will have to first select an emotion category (in bold) and then choose the most appropriate emotion %within the selected category.
%The box at the right of the screen shows at any time the list of emotion categories and associated %emotions.
%If multiple emotions are possible, select the one that in your opinion fits the most.
%After every choice, you will be told if your choice correctly matches the author's to better understand %the task. Press Escape (esc) to hide the notification.
%You must identify 1 emotions or 2 categories correctly to rule out random guessing. Since the task is %ambiguous, we will approve all your submissions if 75\% of your HITs meet the criteria. 
%} 

\subsubsection{Inter-annotator Agreement}
\label{app:MTurkAgreement}

\begin{figure}[!b]
  \centering
  \includegraphics[width=1.0\linewidth]{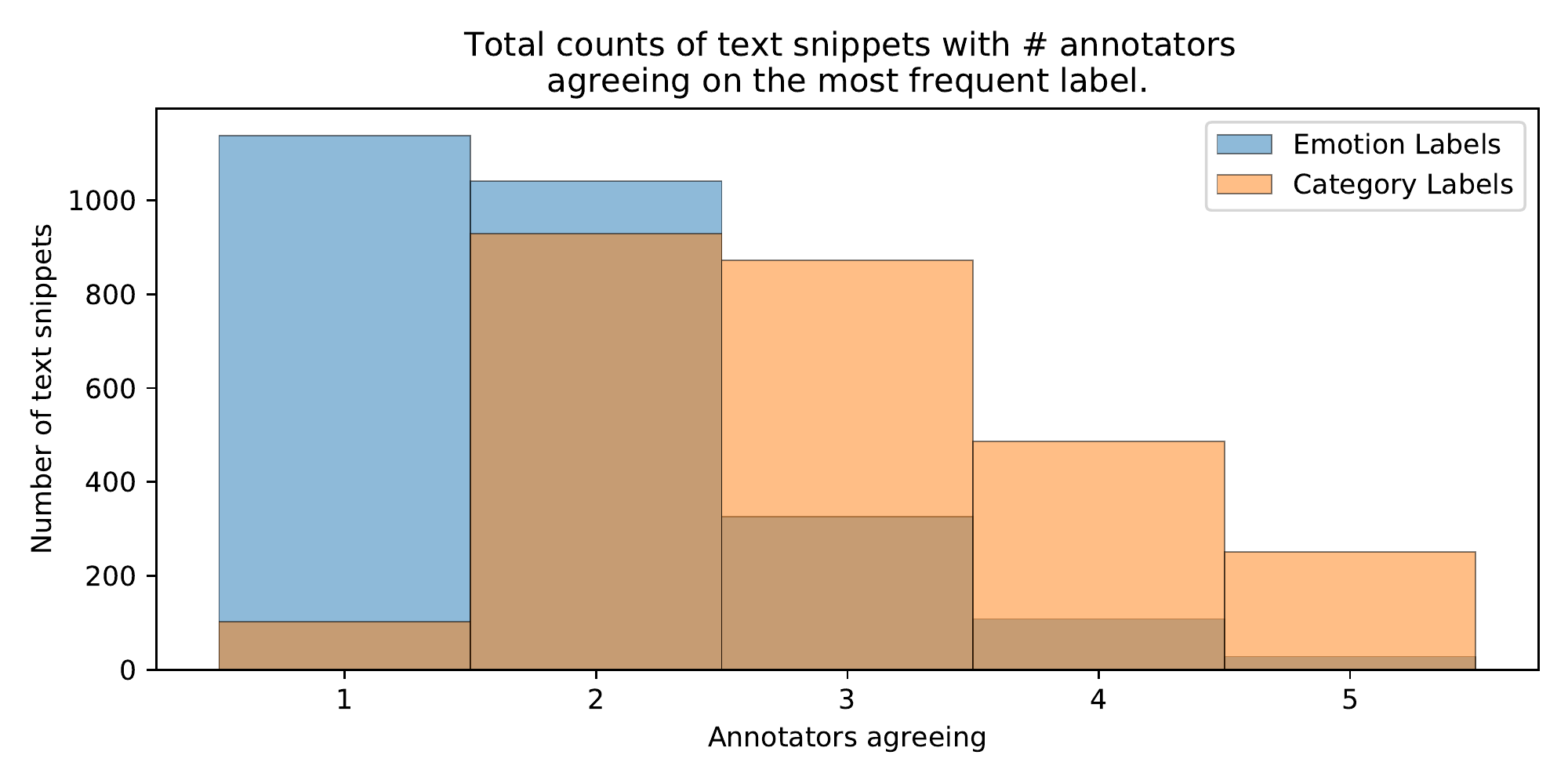}
  \caption{Inter-annotator agreement at the emotion and category levels for the most frequent label. Readers rarely show total disagreement at the category level (average agreement 2.95 $\pm$ 1.03 out of 5), but struggle to exactly pin down the same specific emotions (average 1.81 $\pm$ 0.88 out of 5).}
  \label{fig:MTurkAgreement}
\end{figure}

\Addition{
To validate that the emotion labels in Vent contain meaningful emotional information, we analyze the agreement between annotators (readers). The set of readers performing the annotations varies across snippets, so we focus on the number of readers that assign the same emotion and emotion category to any given snippet. For a given snippet, we consider the reader overlap as the total number of readers that agree with the most frequently assigned label. Since every snippet is shown to 5 different readers, the maximum level of agreement is 5, meaning that all readers agreed on the same judgement, while an agreement of 1 means that each reader chose a different label. The frequency of the overlaps observed across all snippets are shown in~\autoref{fig:MTurkAgreement}. At the emotion level (blue), we find an average agreement of 1.81 ($\pm$ 0.88) readers while at the category level (orange) we find that 2.95 ($\pm$ 1.03) readers agree on average with the most frequent label. These results show that the majority of annotators agree on the category labels, but specific emotions are hard to pin down.
}

\subsubsection{Human vs. Model Performance}

We provide detailed results at the category level of the best performing BERT + Bi-LSTM model and readers in~\autoref{tab:HITModelCategoryResults}
and~\autoref{tab:HITWorkerCategoryResults} respectively. As highlighted in the paper, we find that the precision of the model is lower across all categories while readers show consistently higher recall. On \textcolor[HTML]{DDA0DD}{Affection} and \textcolor[HTML]{800180}{Fear}, readers and the model achieve the same F1-score (0.44 and 0.42 respectively) while otherwise the model outperforms readers in all emotion categories except for \textcolor[HTML]{7DA8D6}{Creativity} and \textcolor[HTML]{00AF02}{Surprise}, which are the worst-performing categories for the model.

\begin{table}[!b]
\centering
\begin{tabular}{@{}lrrrr@{}}
\toprule
\textbf{Emotion} & \textbf{Prec} & \textbf{Rec}  & \textbf{F1}   & \textbf{Sup} \\ \midrule
Affection        & \textbf{0.47} & 0.41          & 0.44          & 270          \\
Anger            & 0.44          & 0.50          & \textbf{0.46} & 270          \\
Creativity       & \textit{0.26} & 0.50          & \textit{0.34} & 300          \\
Fear             & 0.45          & 0.40          & 0.42          & 330          \\
Feelings         & 0.28          & 0.51          & 0.36          & 330          \\
Happiness        & 0.29          & 0.57          & 0.38          & 270          \\
Positivity       & 0.35          & 0.38          & 0.37          & 270          \\
Sadness          & 0.35          & \textbf{0.62} & 0.44          & 330          \\
Surprise         & 0.34          & \textit{0.34} & \textit{0.34} & 270          \\ \bottomrule
\end{tabular}
\caption{Category prediction results for our proposed model, in terms of \textbf{Prec}ision, \textbf{Rec}all, \textbf{F1}-score and \textbf{Sup}port. In \textbf{bold}, the best performing category for each metric; in \emph{cursive}, the worst performing category.}
\label{tab:HITModelCategoryResults}
\end{table}

\begin{table}[!b]
\centering
\begin{tabular}{@{}lrrrr@{}}
\toprule
\textbf{Emotion} & \textbf{Prec} & \textbf{Rec}  & \textbf{F1}   & \textbf{Sup} \\ \midrule
Affection        & \textbf{0.31} & 0.79          & \textbf{0.44} & 270          \\
Anger            & 0.27          & 0.80          & 0.41          & 270          \\
Creativity       & \textit{0.23} & \textit{0.66} & \textit{0.35} & 300          \\
Fear             & 0.30          & \textit{0.66} & 0.42          & 330          \\
Feelings         & \textit{0.23} & 0.70          & 0.35          & 330          \\
Happiness        & 0.25          & 0.69          & 0.37          & 270          \\
Positivity       & \textit{0.23} & 0.70          & \textit{0.34} & 270          \\
Sadness          & 0.27          & \textbf{0.85} & 0.41          & 330          \\
Surprise         & 0.24          & 0.68          & 0.36          & 270          \\ \bottomrule
\end{tabular}
\caption{Category prediction results for the MTurk workers, in terms of \textbf{Prec}ision, \textbf{Rec}all, \textbf{F1}-score and \textbf{Sup}port. In \textbf{bold}, the best performing category for each metric; in \emph{cursive}, the worst performing category.}
\label{tab:HITWorkerCategoryResults}
\end{table}

\begin{figure*}[!ht]
\centering
  \begin{subfigure}{1.0\columnwidth}
  	\centering
    \includegraphics[width=\textwidth]{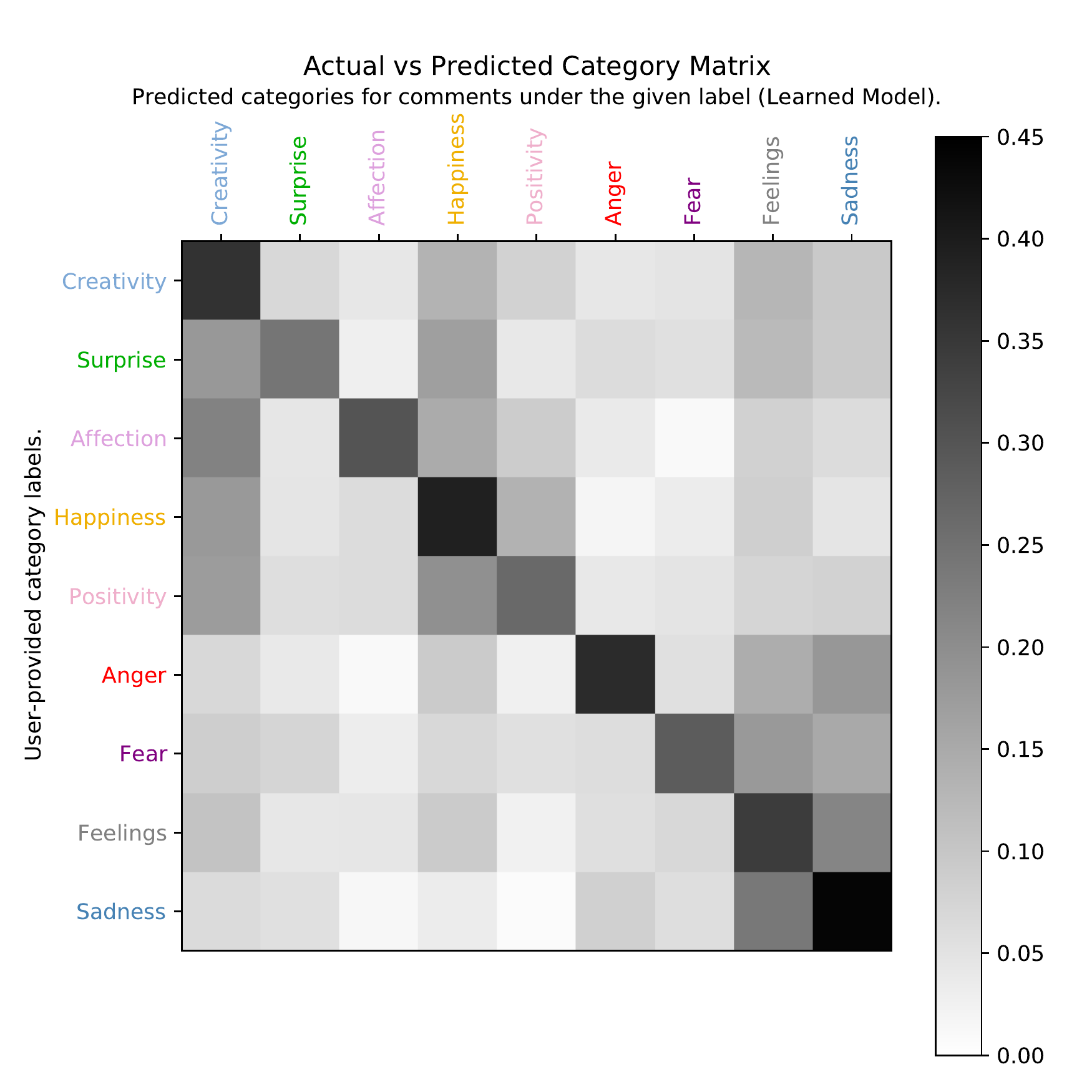}
    \caption{Confusion Matrix for the BERT + Bi-LSTM model.}
  \end{subfigure}
  \begin{subfigure}{1.0\columnwidth}
  	\centering
    \includegraphics[width=\textwidth]{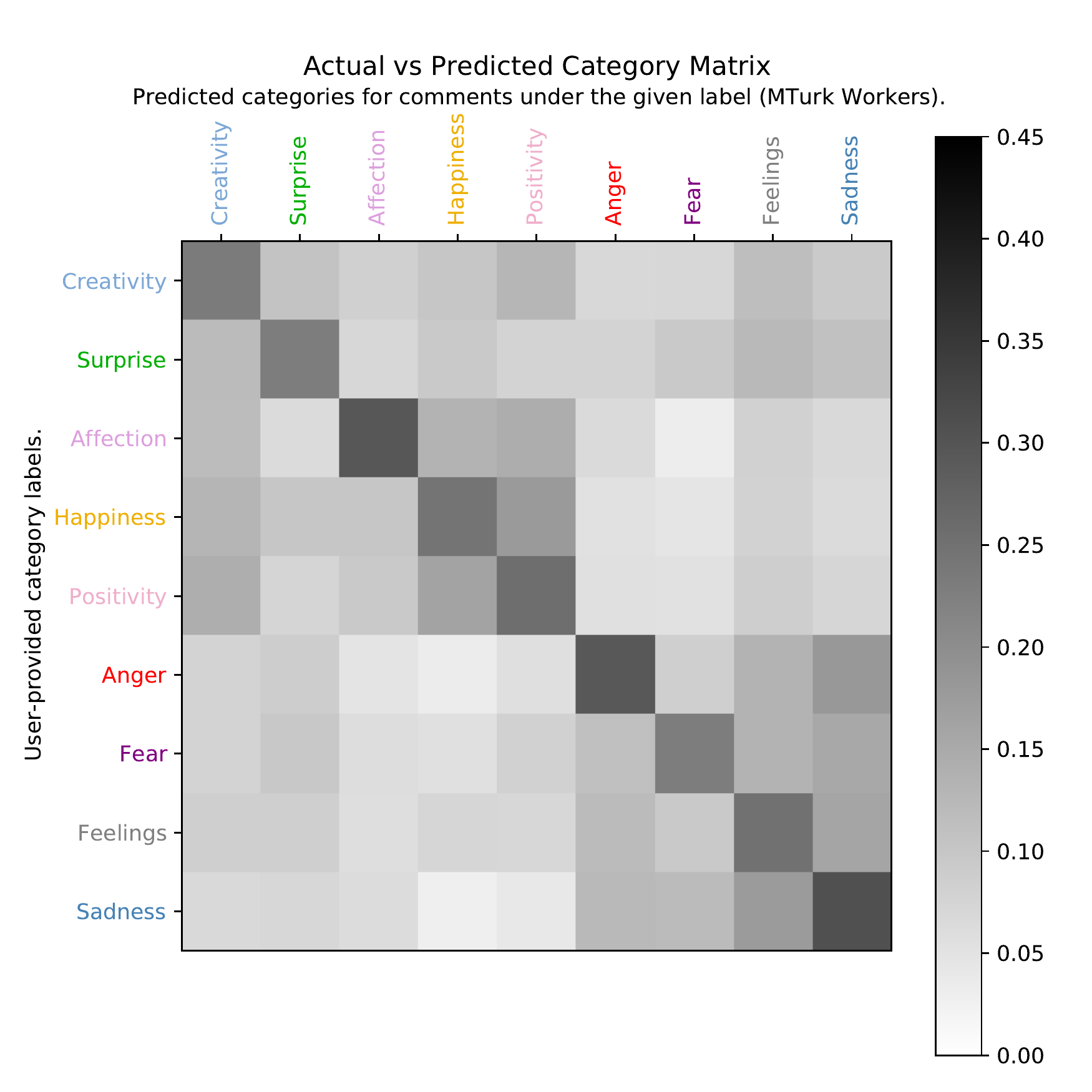}
    \caption{Confusion Matrix for the MTurk Workers.}
  \end{subfigure}
  \caption{Comparison between the confusion matrices on Vent categories computed on the submitted MTurk batch.}
  \label{fig:MTurkConfusionMatrices}
\end{figure*}

We also analyze the confusion matrices between readers and the model (constructed as per the main text). \autoref{fig:MTurkConfusionMatrices} shows that the model is more precisely predicts emotions given a particular author-provided label than the MTurk workers (the diagonal vector is more clearly defined). However, we observe that the confusion between negative (\textcolor[HTML]{FF0000}{Anger}, \textcolor[HTML]{800180}{Fear}, \textcolor[HTML]{808080}{Feelings}, and \textcolor[HTML]{4682B4}{Sadness}) and positive (\textcolor[HTML]{DDA0DD}{Affection}, \textcolor[HTML]{00AF02}{Surprise}, \textcolor[HTML]{7DA8D6}{Creativity}, \textcolor[HTML]{EFAF00}{Happiness}, and \textcolor[HTML]{EFAFCB}{Positivity}) categories is more apparent for workers. 

\begin{figure*}[b]
  \centering
  \includegraphics[width=0.75\linewidth]{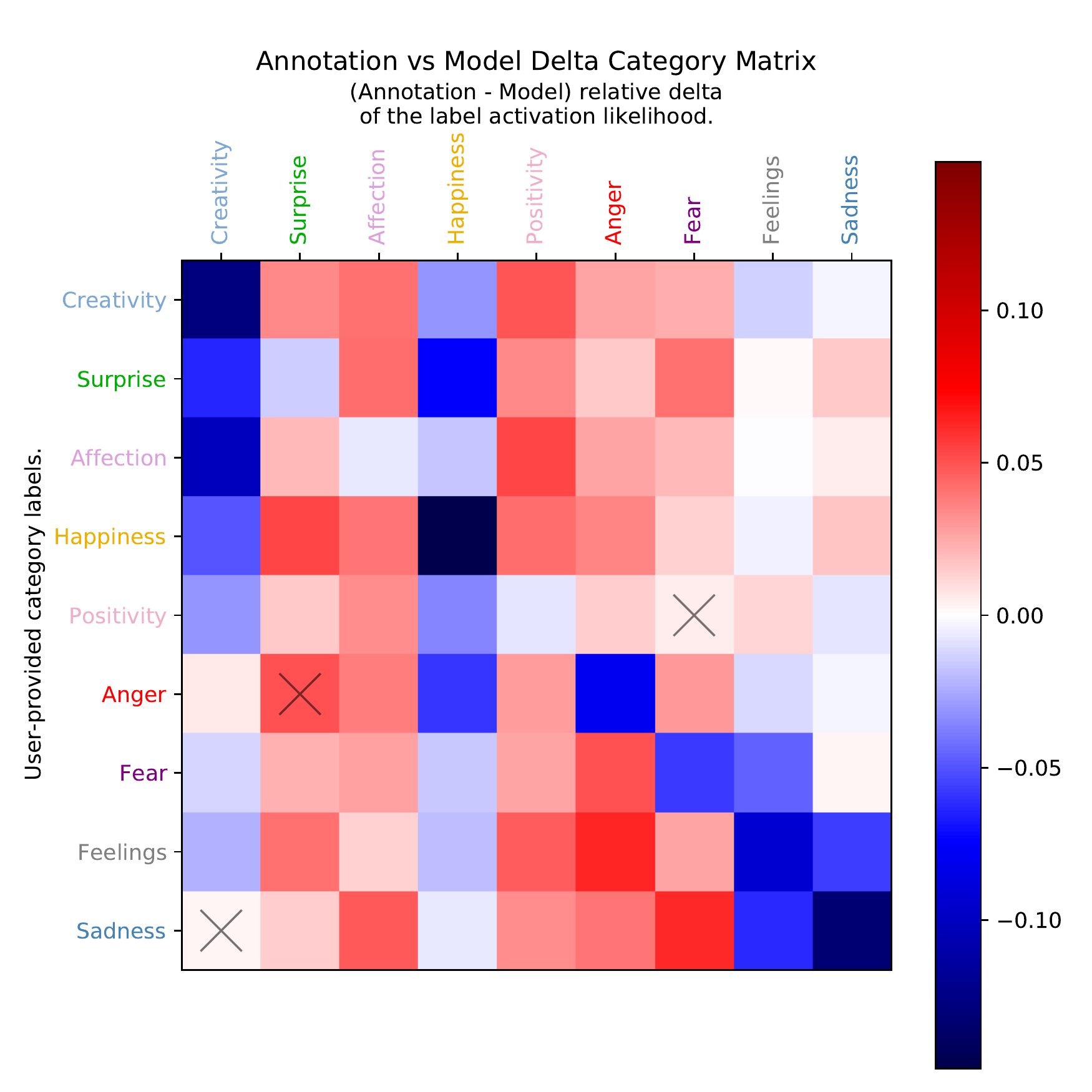}
  \caption{Difference between Model and MTurk worker confusion matrices. Crossed-out cells indicate non-significant differences computed by running bootstrapped simulations on the annotations from the workers ($p = 0.001$).}
  \label{fig:MTurkConfusionMatrixDelta}
\end{figure*}

To further understand the differences between readers and the model, we compute deltas between confusion matrices in~\autoref{fig:MTurkConfusionMatrixDelta}. For an expected $i$-th emotion (row), red cells show when readers are more likely to predict the $j$-th emotion (column) than the model, while blue cells show the opposite case with the model having higher likelihood than readers. To ensure the results are significant, we compute differences over 10,000 bootstrapping runs with the sample size of the evaluation data set (2640 snippets) and perform z-tests on the differences between model and readers.

\autoref{fig:MTurkConfusionMatrixDelta} shows that the model is more likely to output \textcolor[HTML]{7DA8D6}{Creativity} or \textcolor[HTML]{EFAF00}{Happiness}, while workers are more likely to predict \textcolor[HTML]{00AF02}{Surprise}, \textcolor[HTML]{DDA0DD}{Affection} and \textcolor[HTML]{EFAFCB}{Positivity}. We also find blue colors along the diagonals, which agrees with our results on the model having higher precision than the readers.

 % \AKn{Explain what the colors mean in the Figure (in the text and in the caption) and mention the statistical test also in the text. Add how many bootstraps are run}

\subsubsection{Worker Compensation Analysis}

Finally, we discuss the annotator reward of our task, which we defined to ensure that MTurk workers receive fair compensation of at least \$7.25, which is the minimum federal wage in the United States. Upon designing our task and annotation tool, we benchmark ourselves on a sample of 10 HITs using the Requester and Worker Sandbox. In doing so, we identify a minimum amount of time to show feedback after each example (4 seconds), and estimate that every task takes around 90 seconds. 

With our first internal estimate, we schedule a test batch with ~10\% of the data and a reward of \$0.20 per task, which we expect to translate into \$8 per hour. Our test batch helped us identify serious problems with our annotation tool, which allowed workers to submit incomplete tasks. Our acceptance rules flagged their submissions as invalid, which meant that workers could potentially get several rejections. Workers on MTurk depend on a high approval rate to access tasks, so they contacted us and promptly addressed the limitation. 

Additionally, using data from this batch and input from workers, we observed that the task took longer to complete without prior context, and thus our reward was not enough to meet our goal. Given a median amount of time per task of around 130 seconds, we conservatively raise the reward per HIT to \$0.32. This brings our expected payout to \$8.86 per hour, which we use to submit our full batch. 

\begin{figure}[!ht]
  \centering
  \includegraphics[width=1.0\linewidth]{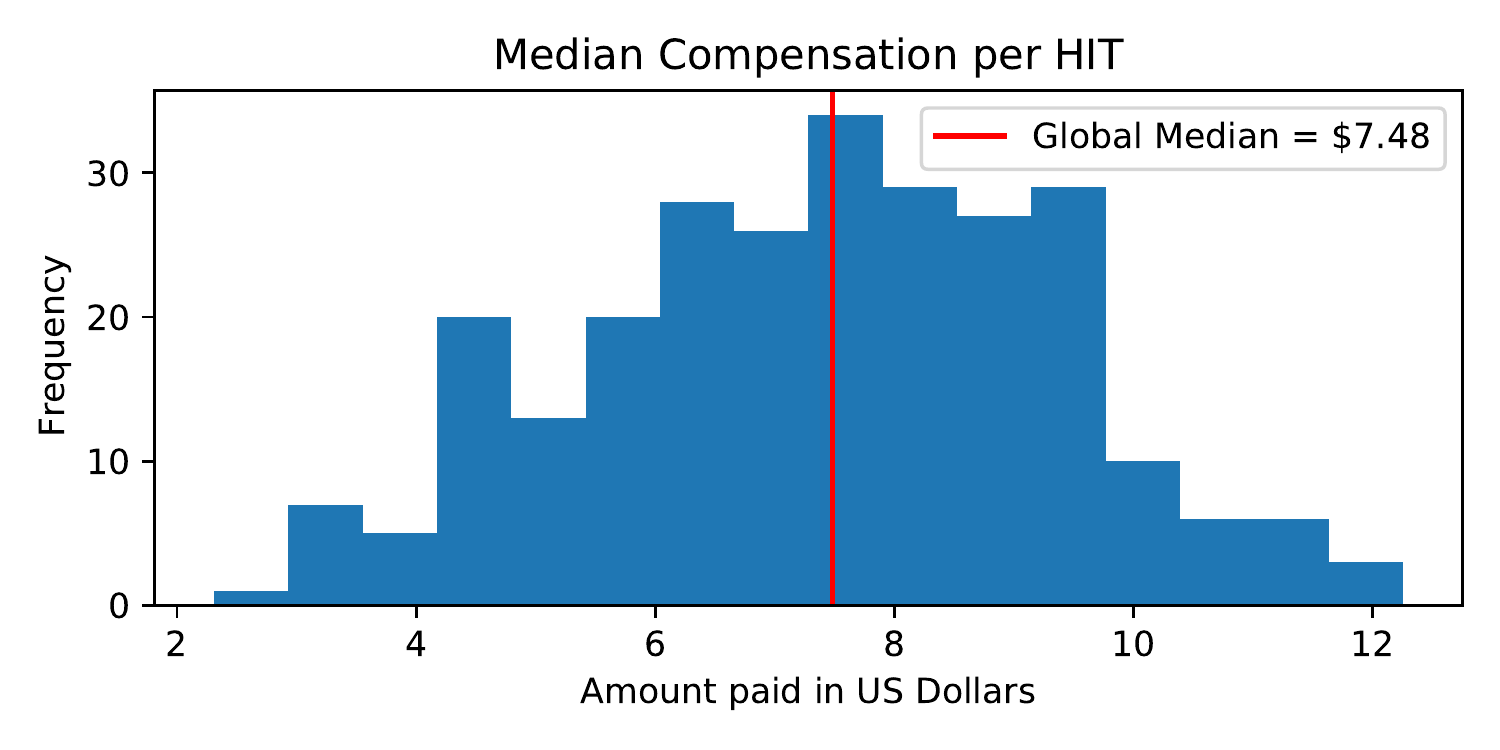}
  \caption{Compensation per HIT. The median compensation per HIT is \$7.48, which means that we met our target of providing the US Federal minimum wage for workers engaging in our task.}
  \label{fig:MTurkCompensation}
\end{figure}

In~\autoref{fig:MTurkCompensation}, we show the distribution of compensations per HIT and the overall median compensation per HIT (red vertical line), which is equivalent to \$7.48 per hour. We notice a large variability in the results, and observe that the final reward per task is lower than expected. This might be caused by the fact that despite text snippets in a HIT being chosen at random, some of the specific HITs remain more challenging than others. Additionally, it might be the case that some workers are more capable of performing the task than others, or that can trade off annotation quality for speed without bypassing our quality checks. Our results show that we achieve our goal of fairly compensating workers per HIT above minimum wage.

\end{appendices}

\end{document}